\documentclass[conference]{IEEEtran}
\usepackage{amssymb}
\usepackage{amsmath}
\usepackage{booktabs} 
\usepackage{multirow}
\usepackage{longtable}
\usepackage{enumitem}
\usepackage{hyperref}
\usepackage{float}

\usepackage[most]{tcolorbox} 

\tcbuselibrary{skins, breakable}
\definecolor{oranBlue}   {HTML}{003A70}   
\definecolor{oranAccent} {HTML}{0072CE}   
\definecolor{oranLight}  {HTML}{E8F2FB}   
\definecolor{oranGold}   {HTML}{F0A500}   
\definecolor{oranGray}   {HTML}{5A6472}   
\definecolor{oranBorder} {HTML}{B0C9E8}   
\definecolor{tagGreen}   {HTML}{1A7F3C}
\definecolor{tagBg}      {HTML}{E6F4EC}
\definecolor{metaBg}     {HTML}{F5F7FA}
\definecolor{valframe}{HTML}{2E8B57}  

\definecolor{lightblue}{RGB}{214, 234, 248}
\definecolor{borderblue}{RGB}{41, 128, 185}
\definecolor{outerbg}{RGB}{248, 249, 250} 
\definecolor{outerborder}{RGB}{170, 170, 170} 
\usepackage[table]{xcolor}
\definecolor{genbg}{HTML}{F0F8FF}     
\definecolor{genframe}{HTML}{2B547E}  
\definecolor{valbg}{HTML}{F5FFFA}     
\definecolor{valframe}{HTML}{2E8B57}  
\definecolor{promptbg}{HTML}{FFFFFF}  
\definecolor{promptframe}{HTML}{808080} 

\newtcolorbox{promptbox}[1]{
  enhanced,
  breakable,
  colback=promptbg,
  colframe=promptframe,
  coltitle=black,
  fonttitle=\bfseries\sffamily,
  title=#1,
  boxrule=0.5pt,
  attach boxed title to top left={yshift=-2.5mm, xshift=4mm},
  boxed title style={colback=promptbg, colframe=promptframe, size=small},
  top=4mm,
  fontupper=\ttfamily\small, 
  left=3mm, right=3mm, bottom=3mm
}

\ifCLASSINFOpdf
\else
\fi

\begin{document}
%
\title{\textbf{TeleEmbedBench}: A Multi-Corpus Embedding Benchmark for Retrieval-Augmented Generation in Telecommunications}



%

\markboth{TeleEmbedBench}{Gajjar {et al.}}

\author{\IEEEauthorblockN{Pranshav Gajjar\IEEEauthorrefmark{1} and
Vijay K Shah\IEEEauthorrefmark{1}}
\IEEEauthorblockA{\IEEEauthorrefmark{1}\textit{NextG Wireless Lab}, North Carolina State University, Raleigh, USA}
}

\title{{TeleEmbedBench: A Multi-Corpus Embedding Benchmark for RAG in Telecommunications}}

\maketitle


%
\IEEEpeerreviewmaketitle

\maketitle
\begin{abstract}
Large language models (LLMs) are increasingly deployed in the telecommunications domain for critical tasks, relying heavily on Retrieval-Augmented Generation (RAG) to adapt general-purpose models to continuously evolving standards. However, a significant gap exists in evaluating the embedding models that power these RAG pipelines, as general-purpose benchmarks fail to capture the dense, acronym-heavy, and highly cross-referential nature of telecommunications corpora. To address this, we introduce \textbf{TeleEmbedBench}, the first large-scale, multi-corpus embedding benchmark designed specifically for telecommunications. The benchmark spans three heterogeneous corpora: O-RAN Alliance specifications, 3GPP release documents, and the srsRAN open-source codebase, comprising 9,000 question-chunk pairs across three standard chunk sizes (512, 1024, and 2048 tokens). To construct this dataset at scale without manual annotation bottlenecks, we employ a novel automated pipeline where one LLM generates specific queries from text chunks and a secondary LLM validates them across strict criteria. We comprehensively evaluate eight embedding models, spanning standard sentence-transformers and LLM-based embedders. Our results demonstrate that LLM-based embedders, such as Qwen3 and EmbeddingGemma, consistently and significantly outperform traditional sentence-transformers in both retrieval accuracy and robustness against cross-domain interference. Additionally, we introduce \textbf{TeleEmbedBench-Clean} to evaluate model robustness against noisy, incomplete user queries. Finally, our analysis reveals that while domain-specific task instructions improve embedder performance for raw source code, they paradoxically degrade retrieval performance for natural language telecommunications specifications.
\end{abstract}

\section{Introduction}

Large language models (LLMs) are increasingly deployed in telecommunications for critical tasks, ranging from network management and RAN optimization to anomaly detection and standards interpretation \cite{zou2025telecomgpt, zhou2024large, lotfi2025oran}. To adapt general-purpose models to this highly specialized domain, the community has primarily relied on two distinct methodologies. fine-tuning \cite{zhang2024scaling, zou2025telecomgpt} and Retrieval-Augmented Generation (RAG) \cite{yilma2025telecomrag, gajjar2025oran}. While fine-tuning updates a model's internal weights using domain-specific prose, producing capable models like TelecomGPT \cite{zou2025telecomgpt}, it is computationally expensive, inherently brittle, and struggles to keep pace with continuously evolving standards. However, fine-tuning is usually considered a superior solution as it is faster for deployment and does not need an additional retrieval step during inference, with RAG being used in conjunction to achieve the maximum possible performance whenever feasible. Yet what the literature consistently reveals is that retrieval is not merely a deployment-time complement to fine-tuning; it is often load-bearing within the fine-tuning process itself. ORANSight-2.0 \cite{gajjar2025oransight} makes this dependency explicit through its RANSTRUCT framework, in which LLM agents retrieve and process O-RAN specification content to synthesize the instruction-tuning dataset used to train the ORANSight models. RAG does not only act as an add-on; it is the mechanism by which domain-grounded training data is produced in the first place. This pattern reflects a broader reality: in a domain where authoritative knowledge is distributed across hundreds of versioned specification documents, retrieval is the only tractable way to surface the precise, localized content that fine-tuning requires. 

At its core, RAG is a specialized data pipeline. A vast library of technical documents, such as 3GPP releases, is chunked and passed through a neural network called the embedding generator \cite{ganiyu2025ai5gtest, yilma2025telecomrag}, which converts text into high-dimensional mathematical vectors that represent the semantic meaning of the content. These embeddings are stored in a vector database, which acts as a searchable long-term memory. When a user query arrives, it is similarly embedded to find the most semantically similar knowledge chunks via similarity search, and those retrieved snippets are injected into the prompt, grounding the LLM with precise domain knowledge. The \textit{embedding model} is the single most critical component of any RAG pipeline, because it determines what information the language model ultimately observes. Despite the central importance of embedding quality, a significant gap remains in the literature: \textit{no rigorous, domain-specific benchmark has been established for evaluating embedding models on telecommunications text}. This gap is particularly acute because telecommunications corpora are distinctly challenging for general-purpose embedders. 

Telecommunications text is dense, acronym-heavy, highly cross-referential, and heterogeneous, spanning formal standards prose from bodies like the O-RAN Alliance and 3GPP alongside semi-structured code and comments from open-source implementations. 
General-purpose benchmarks such as BEIR \cite{thakur2021beir} and MTEB \cite{muennighoff2023mteb}, while valuable, provide limited coverage of technical and standards-heavy domains. Furthermore, the publicly available embedding models are trained on web corpora, and they often struggle with the specialized vocabulary and reference structures characteristic of telecom specifications, as seen in prior work \cite{ganiyu2025ai5gtest} that incorporates Reranker \cite{bge_embedding} based methods to remedy RAG's shortcomings. Recently, some efforts have been made to obtain a Telecom-specific embedding model \cite{gsma2026} \cite{otel2026}, but limited information is available regarding how the training data was obtained. To address this literature gap, we formalize the need for a domain-specific embedding benchmark in telecommunications through \textbf{\textit{TeleEmbedBench}} (\underline{Tele}com \underline{Embed}ding \underline{Bench}mark), a comprehensive multi-corpus embedding benchmark, and the primary contributions of our work are as follows:

\begin{itemize}[leftmargin=*, nosep]

\item \textit{TeleEmbedBench} is the first large-scale benchmark spanning three heterogeneous telecommunications corpora (O-RAN Alliance specifications, 3GPP release documents, and the srsRAN open source codebase), with \textbf{9000} question chunk pairs across three standard chunk sizes.

\item We identify and retain naturally occurring generation artifacts in our main benchmark to simulate noisy, real-world user inputs. Furthermore, we introduce \textit{TeleEmbedBench-Clean}, a filtered subset excluding these artifacts, allowing for an ablation of embedding model robustness against incomplete queries.

\item We conduct a comprehensive evaluation of state-of-the-art eight representative embedding models across chunk sizes and retrieval depths to isolate domain-specific performance and assess robustness to cross-domain interference.

\item We also evaluate domain-specific prompting within embedding models and find that adding instructions paradoxically weakens retrieval, especially for 3GPP and O-RAN specifications. 

\item Both the benchmarks and associated evaluation code have been released publicly\footnote{All resources are available at a centralized HuggingFace Collection: \href{https://huggingface.co/collections/NextGLab/teleembedbench}{HF}.}.

\end{itemize}

The remainder of this paper is organized as follows. Section II provides background on RAG and the taxonomy of embedding model approaches. Section III details the TeleEmbedBench construction methodology, including corpus preparation, question generation, and validation. Section IV describes the evaluation strategy and metrics. Section V presents results across all evaluation strategies and downstream QA experiments. Section VI discusses implications, limitations, and future directions. Section VII surveys related work, and Section VIII concludes.

\section{How do Embedding Models Work?}\label{sec:bg}

\begin{figure}
    \centering
    \includegraphics[width=\linewidth]{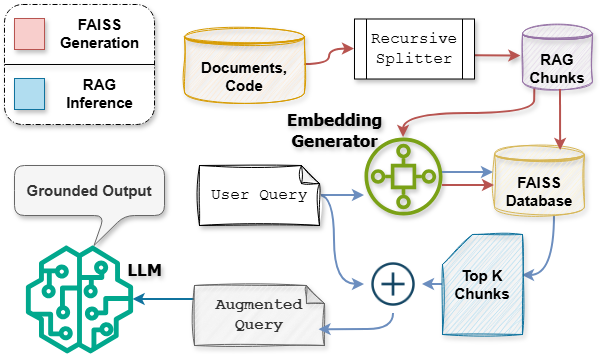}
    \caption{A typical RAG pipeline, from FAISS construction to Inference.}
    \label{fig:rag_demo}
\end{figure}

A typical Retrieval-Augmented Generation (RAG) system operates through a highly scalable \textit{bi-encoder} architecture \cite{reimers2019sentence} divided into two fundamental phases: offline Facebook AI Similarity Search (FAISS) index construction and real-time inference, as illustrated in Figure \ref{fig:rag_demo}. 

During the construction phase, a massive corpus of unstructured knowledge is systematically partitioned into manageable, semantically coherent text chunks. An embedding model then processes these passages, projecting their semantic meaning into high-dimensional, dense numerical vectors that are persistently stored within a specialized vector database like FAISS. Because this encoding happens independently of any user input, the system achieves immense scalability. At inference time, an incoming query is mapped into the \textit{exact} same latent vector space using the identical embedding model. This allows the framework to bypass computationally expensive cross-attention mechanisms, reducing retrieval to a highly efficient nearest-neighbor semantic search to locate the most contextually relevant document vectors. The corresponding textual chunks, thresholded to a specific retrieval rank, are subsequently appended to the original query and fed into an LLM to synthesize a technically grounded response. However, this late-interaction paradigm requires that all rich contextual meaning be \textit{irreversibly} compressed into fixed-size vectors up front. 

Consequently, the embedding model serves as the absolute most critical bottleneck and bridge between the user's intent and the underlying knowledge base, and its compression quality establishes a rigid upper bound on the system's overall efficacy, where any geometric misalignment during the initial semantic search directly precipitates degraded, irrelevant, or entirely fabricated generative outputs \cite{chen2025each}.
These embedding models are trained using contrastive objectives: given a query, the model is rewarded for placing the correct passage nearby in vector space and incorrect passages far away. The difficulty of those incorrect passages called negatives matters enormously. 
Models trained on general web data typically learn to separate passages at a broad topical level, which works well enough when the retrieval task is distinguishing an article about football from one about cooking. It breaks down in telecommunications, where two passages from adjacent clauses of the same specification may look nearly identical on the surface but answer entirely different questions. General-purpose embedders, trained without exposure to this kind of hard negative, tend to fail precisely in the cases that matter most for telecom retrieval. 


We showcase an example of this phenomenon using two adjacent test cases from the NR C-Profile Specification v13:

\begin{figure}[h]
    \centering
    \begin{tcolorbox}[colback=valframe!2!white, colframe=valframe!40!white, 
                      title={\textcolor{black}{\textbf{Clause 6.1.2:} Initial access -- UE Context Creation for Initial Registration}}, 
                      fonttitle=\small\sffamily, fontupper=\small]
    5. The UE responds with an RRCSetupComplete (Registration Request) message.\\
    6. The gNB-DU encapsulates the RRCSetupComplete message in an UL RRC MESSAGE TRANSFER (Registration Request) and sends it to the gNB-CU.\\
    7. The gNB-CU sends an INITIAL UE MESSAGE (Registration Request) to the AMF.\\
    \colorbox{valframe!10}{\parbox{\dimexpr\linewidth-2\fboxsep\relax}{
        \textbf{8. AMF $\rightarrow$ gNB-CU: The AMF responds with an INITIAL CONTEXT SETUP REQUEST.}
    }}\\
    9. The gNB-CU sends a UE CONTEXT SETUP REQUEST to establish the UE context in the gNB-DU. \dots
    \end{tcolorbox}

    \vspace{0.2cm} 

    \begin{tcolorbox}[colback=red!2!white, colframe=red!40!white, 
                      title={\textcolor{black}{\textbf{Clause 6.1.3:} Registration Update Without Follow-on Request}}, 
                      fonttitle=\small\sffamily, fontupper=\small]
    5. The UE responds with an RRCSetupComplete (Registration Request) message.\\
    6. The gNB-DU encapsulates the RRCSetupComplete message in an UL RRC MESSAGE TRANSFER (Registration Request) and sends it to the gNB-CU.\\
    7. The gNB-CU sends an INITIAL UE MESSAGE (Registration Request) to the AMF.\\
    \colorbox{red!10}{\parbox{\dimexpr\linewidth-2\fboxsep\relax}{
        \textbf{8. The AMF sends a DL NAS TRANSPORT (Registration Accept) message.}
    }}\\
    9. The gNB-CU encapsulates the Registration Accept message in a DL RRC MESSAGE TRANSFER \dots
    \end{tcolorbox}
    \caption{An example of a \textit{hard negative}. Steps 5--7 are completely identical across both neighboring clauses, making them nearly indistinguishable to general-purpose embedders despite answering entirely different test cases.}
    \label{fig:hard_negative_example}
\end{figure}

As illustrated in Figure \ref{fig:hard_negative_example}, the initial steps (5 through 7) of both procedures are perfectly identical. If a RAG system processes a highly specific user query such as "\textit{What is the 8th step for the UE Context Creation for Initial Registration test case?}", it must isolate and retrieve the chunk containing Clause 6.1.2. However, because these chunks are stored separately in the vector space, the massive lexical and semantic overlap generated by the identical early steps acts as a trap. A general-purpose embedding model struggles to prioritize the test case title over the dense procedural text, and can potentially score Clause 6.1.3 equally high or higher. Consequently, the retriever fetches the neighboring clause as the context, causing the generation model to confidently output a drastically different and incorrect procedural step.

The embedding models from the available literature broadly fall into two families. 

$\bullet$ The first is what we call the \textit{sentence-transformer} family, which trains a dedicated encoder model typically derived from BERT \cite{devlin2019bertpretrainingdeepbidirectional} or a similar architecture directly on retrieval objectives. These models are compact, fast, and well-understood, but their representational capacity is bounded by the size of the encoder and the diversity of their training data. 

$\bullet$ The second is the \textit{LLM-based embedder}\cite{tao2024llms, zhang2025qwen3} family, which adapts a large pretrained language model to produce passage representations rather than training a retrieval encoder from scratch. Models such as Qwen3-Embedding \cite{zhang2025qwen3} and EmbeddingGemma \cite{vera2025embeddinggemma} fall into this category. Because they inherit their representations from large-scale language model pretraining, they tend to have better coverage of rare and technical vocabulary without any domain-specific training. They also support a useful inference-time technique: a short task instruction can be prepended to the query but not to the document chunks to condition the embedding on the specific retrieval objective. This \textit{asymmetric instruction} strategy can meaningfully improve retrieval without any changes to model weights, and is one of the techniques evaluated in this benchmark.

\section{TeleEmbedBench}

We argue that an effective embedding benchmark for telecommunications must satisfy three key requirements that are not adequately addressed by general-purpose benchmarks such as BEIR\footnote{BEIR is a widely used information retrieval benchmark spanning 18 publicly available datasets across diverse domains, including finance, medicine, and general knowledge sources such as Quora.} \cite{thakur2021beir} and MTEB\footnote{MTEB covers eight embedding tasks across 58 datasets in 112 languages, and includes auxiliary tasks such as clustering and reranking.} \cite{muennighoff2023mteb}, which are as follows.


(I) The benchmark should cover the diverse range of Resources telecom experts actually use. Furthermore, it needs to cover the telecom source codebase and code-mixed dataset\footnote{While \textit{code-mixed} generally implies the usage of two natural languages, here we treat programming code and English text as the two distinct languages}. Finding information in these formats is much harder than searching through regular articles, and a useful benchmark needs to reflect that difficulty. 

(II) It must evaluate robustness across chunking granularities, since RAG deployments vary widely in chunk size, and a foundational model that is brittle to this choice is operationally unreliable, and 

(III) The queries must be faithful proxies for real user behavior, specific enough to have a unique correct answer, and grounded in the technical concepts a domain expert would actually ask about. 

Appropriately addressing these three requirements jointly determines the design of TeleEmbedBench.

\subsection{Corpus Selection}

To cover the heterogeneity of telecommunications retrieval, we construct three sub-corpora. The first draws from O-RAN Alliance specifications: 116 PDF and Word documents covering interface definitions, functional decompositions, and working group outputs \cite{gajjar2025oran}. The second consists of all 3GPP Release~19 documents in DOCX, DOC, and PDF formats. The third is the srsRAN open-source 5G codebase~\cite{barbosa2025open} version \texttt{25.10.0}, comprising C++ source files, header files, and inline documentation in Markdown and YAML.


\subsection{Document Processing Pipeline}
\begin{figure}
    \centering
    \includegraphics[width=\linewidth]{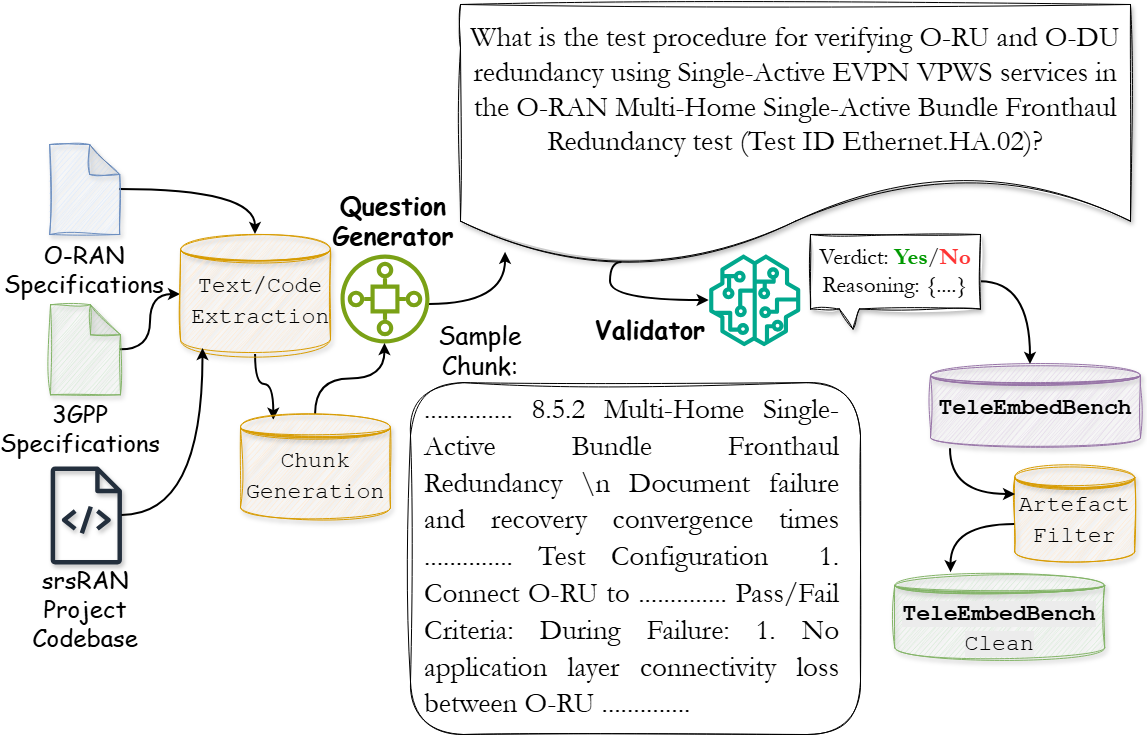}
    \caption{High-level overview of the TeleEmbedBench question-chunk pair generation pipeline}
    \label{fig:mtele-pipeline}
\end{figure}

All three sub-corpora pass through a unified processing pipeline with format-specific adaptations at the extraction stage, which broadly consists of PDFs, Docs, and srsRAN source files, which are scattered across multiple formats. We restrict the ingestion of tables and figures before chunking and benchmark only the core textual and programmatic content, avoiding the current limitations and complexities of multimodal parsing. 

Chunking is performed using LangChain's \texttt{RecursiveCharacterTextSplitter} \cite{mavroudis2024langchain} with the \texttt{cl100k\_base} tokenizer \cite{islam2025gpt} at three target sizes: \textit{512}, \textit{1024}, and \textit{2048} tokens with a 50-token overlap at boundaries to preserve context. Chunks falling outside a 50\%-150\% tolerance window around the target size are discarded to avoid degenerate segments. The separator hierarchy is adapted by domain, for natural-language corpora split on paragraph and sentence boundaries, while the srsRAN corpus splits on code-structure delimiters such as closing braces, semicolons, and blank lines, to avoid bisecting function bodies. The three chunk sizes are central to the benchmark's design rather than incidental. RAG deployments span a wide range of operating points, from short precision-oriented windows in latency-constrained settings to broad context windows where answers span multiple sentences. An embedding model should be robust across this range. Evaluating at all three sizes lets TeleEmbedBench characterize how model performance degrades or holds as granularity changes, a property that single-size benchmarks cannot measure.

\subsection{Question-Chunk Pair Generation}

Another challenge in building such a retrieval benchmark is scale, as for each query, a designated correct chunk must be identified, and producing thousands of such pairs manually with the domain expertise that telecommunications requires is not tractable. The question is whether this annotation burden can be eliminated without sacrificing the faithfulness of the resulting queries. Our approach rests on a single core insight:

\begin{tcolorbox}[
    colback=gray!8,
    colframe=black,
    boxrule=0.6pt,
    arc=2pt,
    left=3pt,
    right=3pt,
    top=3pt,
    bottom=3pt,
    width=\linewidth
]
\textit{If a question is generated by conditioning a language model on a specific text chunk, then that \textbf{chunk is the uniquely correct retrieval target} for that question, and a well-functioning embedding model should rank it first.}
\end{tcolorbox}

\noindent This reframes benchmark construction as a generation problem rather than an annotation problem. Instead of asking human experts to write queries and then label which chunk answers them, we invert the process: derive questions from chunks, so that the ground-truth retrieval target is known by construction. By conditioning generation on the chunk, we recover that specificity without manual effort. A high-level overview of the approach is shown in Figure~\ref{fig:mtele-pipeline}, and as it is shown, the Generation is performed by a Question Generator Agent which uses a \texttt{Qwen3:14b}~\cite{yang2025qwen3}. This agent receives the chunk text and is instructed with the prompt template as shown in Appendix \ref{prompts} Figure~\ref{fig:prompts-generator} to produce a single, conversational, technically grounded question that is strictly answerable from the chunk alone and formulated such that the source chunk constitutes the uniquely correct top retrieval result. Generic or cross-chunk questions are explicitly prohibited in the system prompt. 

As Generation alone is insufficient and a prompted model will occasionally produce questions that are too broad to discriminate the target chunk from its neighbors, or too vague to have a unique answer. We therefore adopt a two-model pipeline inspired from the ORANBench paper \cite{gajjar2025oran}, and assess each generated pair by a \texttt{Gemma3:12b}~\cite{gemmateam2025gemma3technicalreport} which acts as a gatekeeper across four criteria whether the chunk contains substantive text rather than tabular or non-textual content; whether the question is clear, specific, and well-formed; whether the question is directly answerable from the chunk without requiring external context; and whether the source chunk would constitute the top retrieval result for the question. The detailed prompt for the Validator is available in Appendix \ref{prompts} Figure~\ref{fig:prompts-validator}. A pair must pass all four criteria to be retained; failure on any single dimension results in discarding the pair. Using distinct models for generation and validation reduces the risk that generator-specific biases systematically inflate acceptance rates.

Generation iterates over a shuffled pool of chunks until 1,000 validated pairs are collected per chunk size per sub-corpus, yielding 9,000 pairs in total. Each record stores the chunk text, the generated question, the source document identifier, the chunk size in tokens, and the validator's recorded reasoning, which is presented in the Appendix \ref{appx:val}. This central insight that generation from chunks substitutes for manual annotation is a hypothesis rather than an assumption. A generated question could, in principle, be so generic that many chunks would serve equally well as its answer, breaking the one-to-one correspondence that the benchmark depends on. Section~\ref{sec:downstream} tests this directly, by measuring whether retrieval accuracy on TeleEmbedBench predicts end-to-end answer quality on ORANBench. A strong correlation between intrinsic retrieval performance and downstream QA accuracy constitutes direct evidence that LLM-generated questions are faithful proxies for the queries that real practitioners pose to a telecom RAG system.

\subsection{Generation Artifacts and TeleEmbedBench-Clean}
\begin{figure}[htbp]
    \centering
    
    \begin{tcolorbox}[colback=outerbg, colframe=outerborder, boxrule=1pt, arc=6pt, left=12pt, right=12pt, top=12pt, bottom=12pt]
        
        \textbf{Question:} \\
        What are the key 3GPP and O-RAN technical specifications that define the architecture, interfaces, and general principles for E-UTRAN, NG-RAN, and O-RAN components such as X2, Xn, E1, F1, A1
        
        \vspace{1em}
        \hrule
        \vspace{1em}
        
        \textbf{Source Document:} O-RAN.WG1.OAD-R003-v11.00 (2).docx \\
        \textbf{Chunk Size:} \texttt{512}

        \begin{tcolorbox}[colback=lightblue, colframe=borderblue, boxrule=1pt, arc=4pt, title=\textbf{Validation Reasoning}]
        The question is clear and directly answerable from the provided chunk. The chunk lists numerous relevant technical specifications, making it a suitable retrieval target for this question. The question is specific and well-formed, targeting a defined set of specifications.
        \end{tcolorbox}
        
    \end{tcolorbox}
    
    \caption{Document Retrieval Validation and Metadata}
    \label{fig:validation_box}
\end{figure}

During the automated generation of the 9,000 question-answer pairs, we observe naturally occurring generation artifacts where the \texttt{Qwen3:14b} occasionally truncated questions mid-sentence. We formally define these artifacts as questions that do not end with a terminal punctuation mark as they lack a `?' or `.' as seen in figure \ref{fig:validation_box}. Despite being incomplete, these queries successfully passed our validation pipeline because the partial semantic signal was still sufficient for the validator to uniquely identify the source chunk. Table \ref{tab:artifacts} details the distribution of these 1,640 truncated questions across the three corpora and chunk sizes.

\begin{table}[ht]
\centering
\caption{Distribution of Generation Artifacts (Truncated Questions)}
\label{tab:artifacts}
\begin{tabular}{lcccc}
\toprule
\multirow{2}{*}{\textbf{Corpus}} 
  & \multicolumn{3}{c}{\textbf{Chunk Size (tokens)}} 
  & \multirow{2}{*}{\textbf{Total}} \\
\cmidrule(lr){2-4}
 & \textbf{512} & \textbf{1024} & \textbf{2048} & \\
\midrule
O-RAN Specifications   & 153 & 180 & 224 & 557 \\
3GPP Specifications   & 145 & 212 & 227 & 584 \\
srsRAN Project & 151 & 164 & 184 & 499 \\
\midrule
\textbf{Total} & \textbf{449} & \textbf{556} & \textbf{635} & \textbf{1,640} \\
\bottomrule
\end{tabular}
\end{table}

Rather than discarding these samples, we retain them in the main TeleEmbedBench dataset. In real-world Retrieval-Augmented Generation (RAG) applications, user queries are frequently noisy, informal, or incomplete. Recent literature emphasizes the critical need for embedding models to remain robust against such query-level perturbations and typographical noise \cite{campos2023noise, sidiropoulos2025improving}. By keeping these artifacts, the main benchmark serves as a realistic testbed for evaluating model robustness. Concurrently, we introduce \textbf{TeleEmbedBench-Clean}, a subset comprising the remaining 7,360 cleanly formatted questions. This provides a controlled environment to isolate and quantify the impact of query noise on retrieval performance.

\subsection{Benchmark Statistics}
\begin{table}[ht]
\centering
\caption{Total Chunks per Corpus and Chunk Size}
\label{tab:chunks}
\begin{tabular}{lccccc}
\toprule
\multirow{2}{*}{\textbf{Corpus}} 
  & \multicolumn{3}{c}{\textbf{Chunk Size (tokens)}} 
  & \multirow{2}{*}{\textbf{Total}} \\
\cmidrule(lr){2-4}
 & \textbf{512} & \textbf{1024} & \textbf{2048} & \\
\midrule
O-RAN Specifications   &   9,052 &  4,802 & 2,297  &  16,151 \\
3GPP Specifications   & 181,544 & 86,388 & 42,131 & 310,063 \\
srsRAN Project &  19,163 &  9,363 & 4,491  &  33,017 \\
\midrule
\textbf{Total} & \textbf{209,759} & \textbf{100,553} & \textbf{48,919} & \textbf{359,231} \\
\bottomrule
\end{tabular}
\end{table}

Table \ref{tab:chunks} summarizes the resulting corpus statistics. Despite the 3GPP corpus dominating by volume, TeleEmbedBench samples question-chunk pairs uniformly across all nine (corpus, chunk size) configurations rather than proportionally. This is a deliberate design choice: proportional sampling would effectively collapse the benchmark into a 3GPP-only evaluation, allowing strong performance on that corpus to mask failures on srsRAN or O-RAN.

\section{Evaluation Strategy}

\subsection{Metrics}
Our primary metric is the \textbf{Retrieval Accuracy}, which measures the fraction of queries for which the ground-truth chunk appears as the absolute first retrieved result. Mathematically, given a set of $N$ queries, let $r_i$ denote the rank of the ground-truth chunk for the $i$-th query. The Retrieval accuracy is defined as:

\begin{equation}
\text{Retrieval Accuracy} = \frac{1}{N} \sum_{i=1}^{N} \mathbb{I}(r_i = 1)
\end{equation}

where $\mathbb{I}(\cdot)$ is the indicator function that evaluates to $1$ if the ground-truth chunk is correctly placed at rank 1, and $0$ otherwise. As a secondary ranking metric, we report the \textbf{Mean Reciprocal Rank (MRR)} \cite{wu2011optimizing}, which rewards models that place the correct chunk higher up in the ranked list. In this formulation, if the correct chunk is found within the top 5, we take the inverse of its rank; if it falls outside the top 5, it contributes $0$ to the sum:

\begin{equation}
\text{MRR} = \frac{1}{N} \sum_{i=1}^{N} 
\begin{cases} 
\frac{1}{r_i} & \text{if } r_i \le 5 \\ 
0 & \text{otherwise} 
\end{cases}
\end{equation}

Finally, on the efficiency axis, we record the \textbf{inference latency} in \textit{ms} (denoted as $L_{\text{inf}}$) to support the accuracy/efficiency trade-off analysis.

\subsection{FAISS construction}
Now, for each combination of embedding model, chunk size, and sub-corpus, we construct a separate FAISS index~\cite{douze2025faiss}, which has been the go-to database for multiple telecom RAG systems \cite{ganiyu2025ai5gtest, gajjar2025oran, gajjar2025oransight}. We define three complementary strategies for evaluating FAISS indices of a particular chunk size that together cover the range of real deployment scenarios. 

$\bullet$ \textbf{Strategy~A (Per-Corpus).} Each sub-corpus is evaluated independently using its own FAISS index. This isolates domain-specific retrieval performance. 

$\bullet$ \textbf{Strategy~B (Macro-Averaged).} Top-$1$ accuracy averaged uniformly across all three sub-corpora. This yields a single scalar ranking per (model~$\times$~chunk size) configuration.

$\bullet$ \textbf{Strategy~C (Merged FAISS).} A single index is built over all three corpora combined, and every query is evaluated against this unified knowledge base. This strategy simulates a realistic production deployment where a single vector store spans heterogeneous document types, and measures the degree to which cross-domain interference degrades per-corpus accuracy relative to the strategy~A baseline. We leverage the same strategies for the main benchmark and the clean subset.

\subsection{Downstream Analysis}
Finally, to evaluate the benchmark generation method itself, we conduct a downstream analysis in Section~\ref{sec:downstream}. We primarily use the ORANBench dataset, as we leverage the same 116 documents from the benchmark's paper in our work to construct the FAISS indices\footnote{For further discussion regarding the necessity of exact document alignment between the benchmark and the underlying vector database, refer to Section \ref{limitations}.}. We use two flagship LLM models, Gemma3:4B and Gemma3:27B, and replicate the RAG inference pipeline from the ORANSight implementation, as shown in the paper \cite{gajjar2025oran}.

\subsection{Embedding Models}
\begin{table}[t]
\centering
\caption{Embedding models evaluated in TeleEmbedBench. Here, PRMs indicate the model parameters.}
\label{tab:models}
\scriptsize
\renewcommand{\arraystretch}{0.9}
\resizebox{\columnwidth}{!}{%
\begin{tabular}{lll}
\toprule
\textbf{Full Model Name} & \textbf{Alias} & \textbf{PRMs} \\
\midrule
\texttt{all-MiniLM-L6-v2} \cite{yin2024study} & MiniLM & 22\,M  \\
\texttt{bge-small-en-v1.5} \cite{bge_embedding}                 & BGE-S  & 33\,M  \\
\texttt{all-mpnet-base-v2} \cite{siino2024all} & MPNet  & 110\,M \\
\texttt{bge-base-en-v1.5} \cite{bge_embedding}                  & BGE-M  & 109\,M \\
\texttt{paraphrase-multilingual-mpnet-base-v2} \cite{reimers2019sentence} & Multi-MPNet & 278\,M \\
\texttt{bge-large-en-v1.5} \cite{bge_embedding}                 & BGE-L  & 335\,M \\
\rowcolor{gray!20} \texttt{EmbeddingGemma-300m} \cite{vera2025embeddinggemma}  & EGemma & 300\,M \\
\rowcolor{gray!20} \texttt{Qwen3Embedding} \cite{zhang2025qwen3} & Qwen3  & 600\,M \\
\bottomrule
\end{tabular}%
}
\end{table}

Table~\ref{tab:models} lists the eight embedding models evaluated in this benchmark, covering both major architectural families described in Section~\ref{sec:bg}. The sentence-transformer group spans four parameter scales, and the LLM-based embedder group consists of Qwen3 and EGemma, both of which support task-instruction prefixes at inference time. We also evaluate the effect of telecom-specific prompts that leverage the instruction tuning in section V \ref{sec:instruction}

\section{Results}
\subsection{TeleEmbedBench}
\subsubsection{Strategy A (Per-Corpus)}
\begin{figure}[t]
    \centering
    \includegraphics[width=\columnwidth]{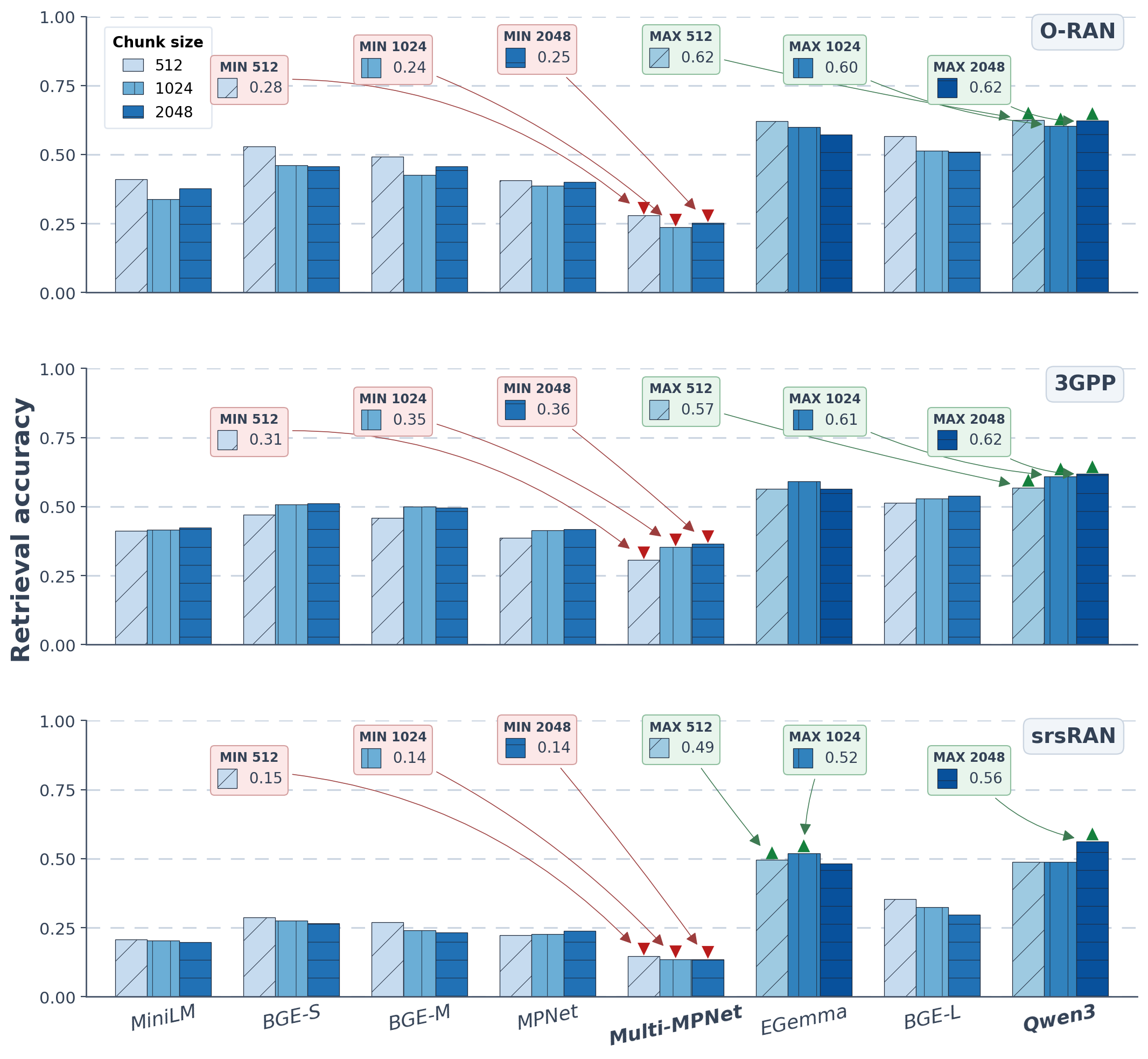}
    \caption{Top-1 retrieval accuracy under Strategy A (per-corpus evaluation) across three sub-corpora. Each group of bars corresponds to one embedding model. Green triangles ($\blacktriangle$) indicate the highest retrieval accuracy model at each chunk size; red triangles ($\blacktriangledown$) indicate the lowest.}
    \label{fig:Strategy_a}
\end{figure}

Figure~\ref{fig:Strategy_a} shows that the majority of embedding models fail to exceed 0.50 top-1 retrieval accuracy even on the natural-language corpora. Of the eight models evaluated, only the LLM-based embedders Qwen3 and EGemma consistently breach this threshold, with Qwen3 peaking at \textbf{0.625} on O-RAN and \textbf{0.619} on 3GPP, and EGemma closely following at 0.621 and 0.592, respectively. The sentence-transformer family clusters well below, with MiniLM and MPNet remaining in the 0.38-0.41 range and BGE-S and BGE-M converging around 0.42-0.53 despite a 3$\times$ difference in parameter count. Only BGE-L at 335M parameters approaches competitive performance, reaching 0.566 on O-RAN and 0.539 on 3GPP, yet it still trails EGemma by 0.06-0.08 points while requiring over 15$\times$ the parameters of MiniLM for a 0.15 gain, pointing to a representational ceiling inherent to the sentence-transformer architecture rather than a gap closable by scale. Multi-MPNet, the second largest sentence-transformer at 278M parameters, inverts this trend entirely, performing worst in every configuration, which we believe is due to the multilingual training objective interfering with the fine-grained semantic distinctions that telecommunications retrieval demands. On the srsRAN code corpus, the performance gap widens further, with BGE-L falling to 0.354 at 512 tokens and declining to 0.297 at 2048, and Multi-MPNet reaching a minimum of 0.135. Even the LLM-based embedders degrade on code, though Qwen3 uniquely benefits from larger chunks, rising from 0.489 at 512 and 1024 tokens to 0.563 at 2048 and overtaking EGemma, which peaks at 1024 (0.520) before retreating to 0.482, a divergence that suggests exploiting broader syntactic context in code requires a representational capacity absent in smaller architectures.

Furthermore, analyzing the impact of chunk sizes across the different corpora, we can notice that for O-RAN and srsRAN, the 512-token chunk size usually performs better, likely because smaller chunks capture the localized, specific nature of the queries more effectively in these domains. This trend changes for 3GPP, where the 2048-token chunk size demonstrates superior performance, indicating that the dense and highly cross-referential nature of 3GPP specifications benefits from broader contextual windows during retrieval. 

\subsubsection{Strategy B (Macro-Averaged)}

\begin{figure}[t]
    \centering
    \includegraphics[width=\columnwidth]{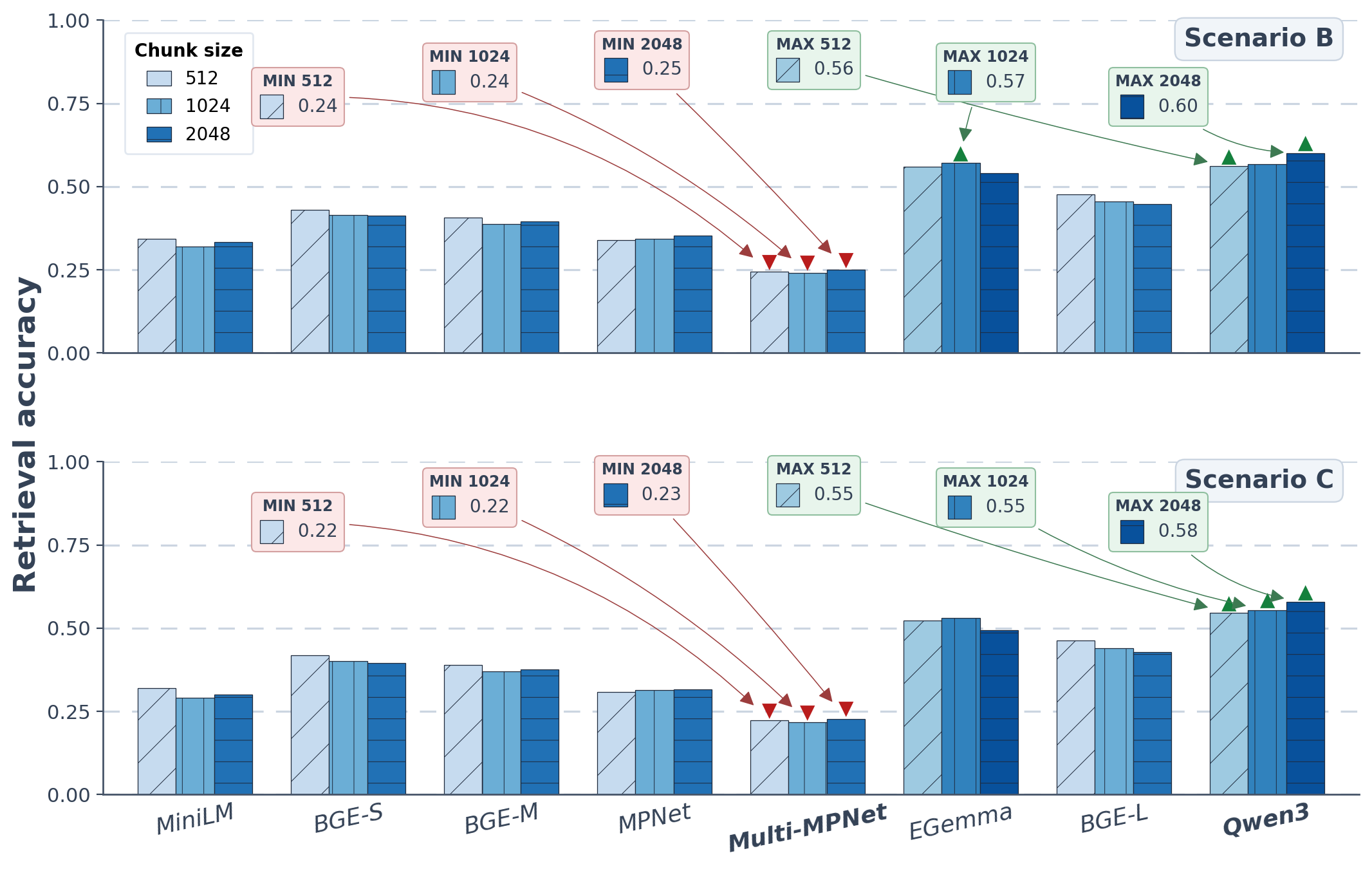}
    \caption{Top-1 retrieval accuracy under Strategy B (Macro-Averaged) and Strategy C (Merged FAISS). Each group of bars corresponds to one embedding model. Green triangles ($\blacktriangle$) indicate the highest retrieval accuracy model at each chunk size; red triangles ($\blacktriangledown$) indicate the lowest.}
    \label{fig:Strategy_bc_acc}
\end{figure}

\begin{figure}[t]
    \centering
    \includegraphics[width=\columnwidth]{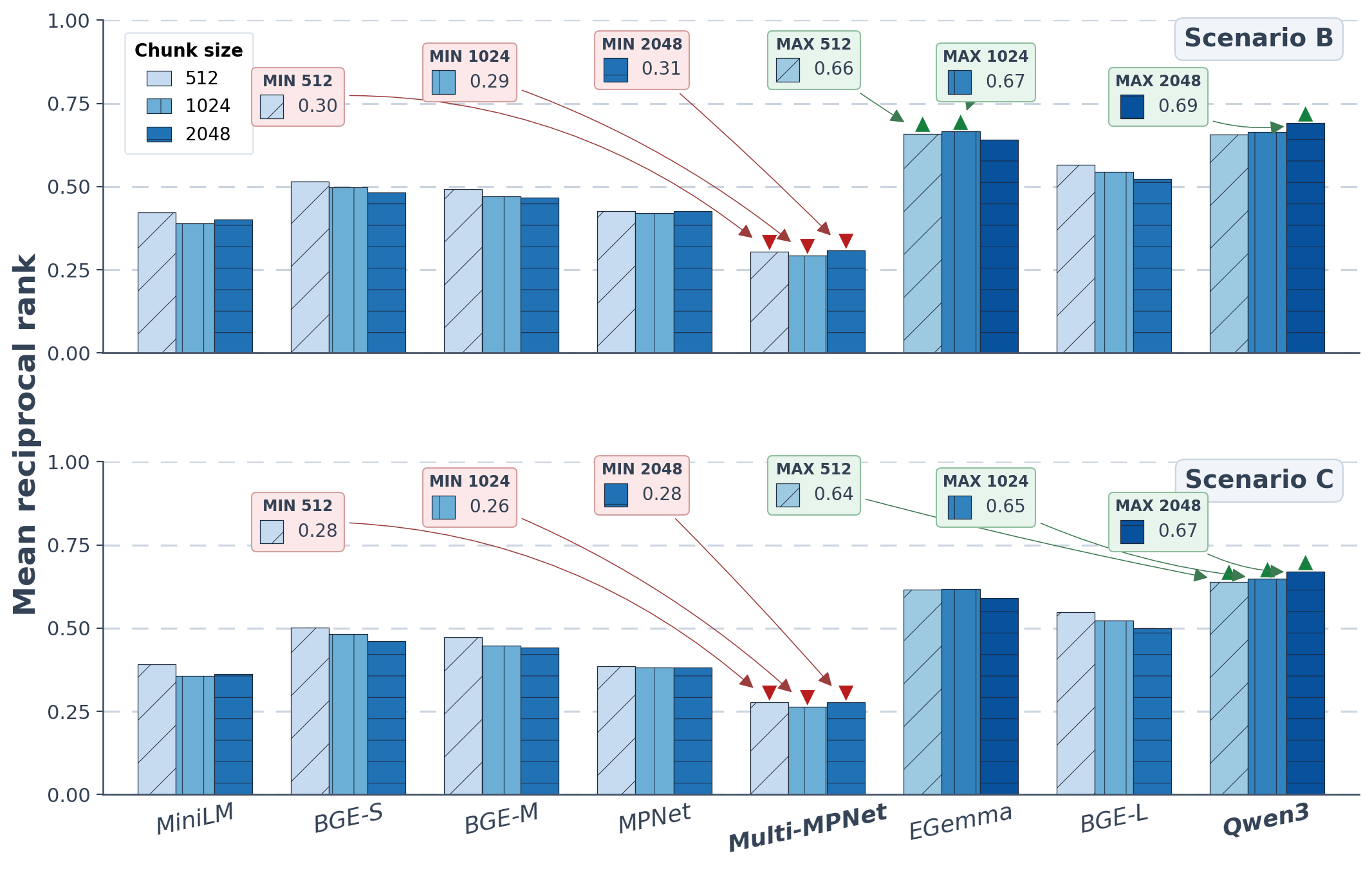}
    \caption{MMR under Strategy B (Macro-Averaged) and Strategy C (Merged FAISS). Each group of bars corresponds to one embedding model. Green triangles ($\blacktriangle$) indicate the highest retrieval accuracy model at each chunk size; red triangles ($\blacktriangledown$) indicate the lowest.}
    \label{fig:Strategy_bc_mrr}
\end{figure}

Figure~\ref{fig:Strategy_bc_acc} (top panel) illustrates the macro-averaged top-1 retrieval accuracy across all three corpora under Strategy B. Consistent with the per-corpus results, the LLM-based embedders Qwen3 and EGemma maintain a significant lead over the sentence-transformer models. Qwen3 achieves the highest macro-averaged accuracy, particularly excelling at the 2048-token chunk size, while EGemma shows strong performance across all sizes but peaks at 1024 tokens. Multi-MPNet consistently exhibits the lowest performance across all configurations, underscoring its unsuitability for this domain. The remaining sentence-transformers (MiniLM, BGE-S, BGE-M, MPNet) cluster together with marginal differences, reaffirming the representational ceiling observed in Strategy A. Figure~\ref{fig:Strategy_bc_mrr} (top panel) corroborates these findings using Mean Reciprocal Rank (MRR), demonstrating that even when models fail to place the correct chunk at rank 1, Qwen3 and EGemma are significantly more effective at surfacing it within the top 5 results compared to smaller architectures.

\subsubsection{Strategy C (Merged FAISS)}

Under Strategy C, we observe a slight degradation in overall retrieval accuracy due to cross-domain interference. As shown in the bottom panel of Figure~\ref{fig:Strategy_bc_acc}, despite the increased complexity of the search space, the relative ranking of the models remains largely stable. Qwen3 and EGemma continue to dominate, successfully distinguishing between visually similar but semantically distinct passages across different standards and codebases. The robustness of these LLM-based embedders is further highlighted in the MRR results (Figure~\ref{fig:Strategy_bc_mrr}, bottom panel), where they sustain strong ranking performance despite the unified index. Conversely, the smaller sentence-transformer models suffer more noticeable declines, indicating that their representations are less capable of cleanly separating O-RAN, 3GPP, and srsRAN contexts when queried simultaneously. This demonstrates that larger representational capacity is essential not just for domain-specific precision, but also for resisting cross-corpus collision in realistic, unified vector databases.

\subsection{TeleEmbedBench-Clean}
\subsubsection{Strategy A (Per-Corpus)}

\begin{figure}[t]
    \centering
    \includegraphics[width=\columnwidth]{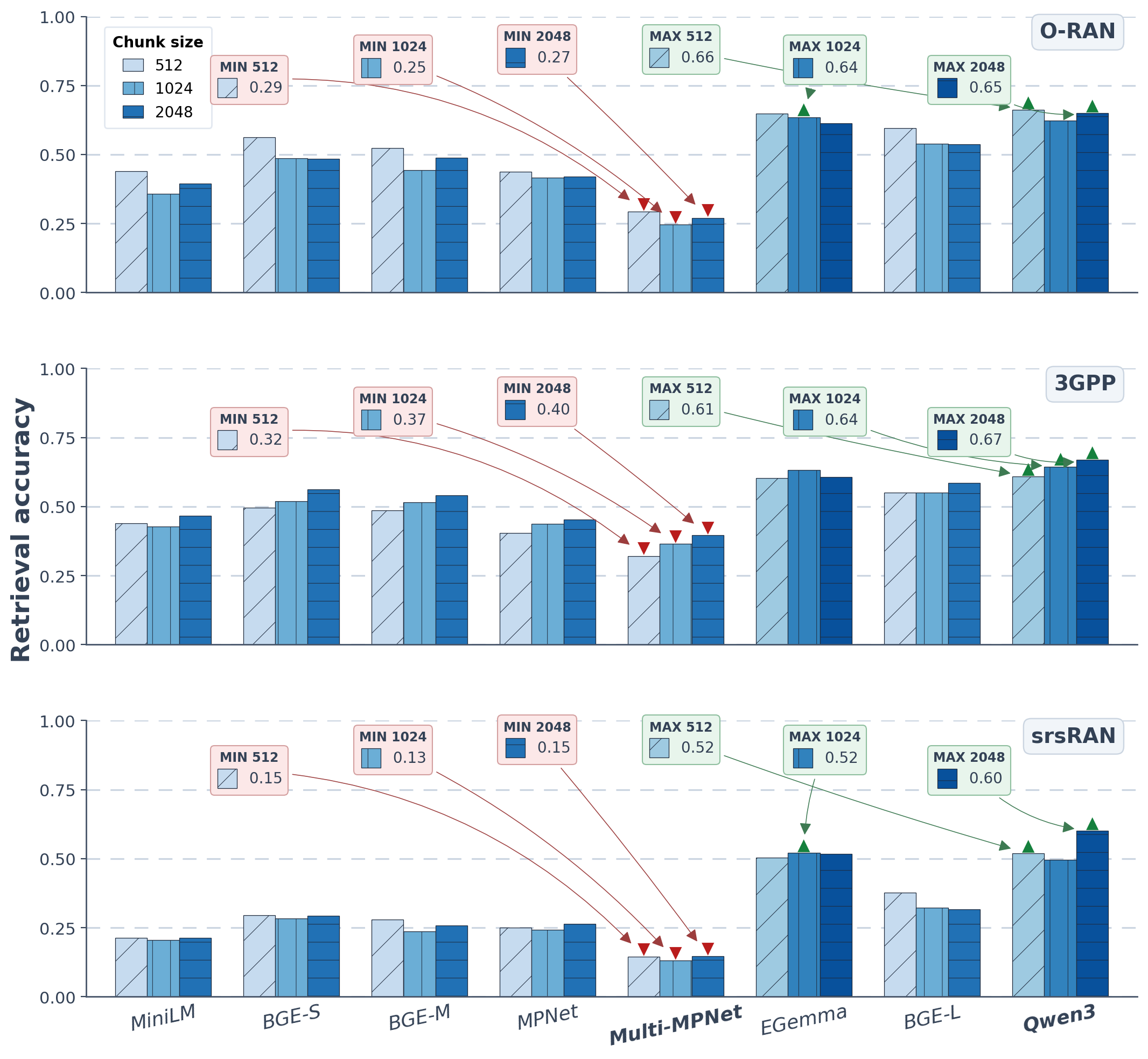}
    \caption{Top-1 retrieval accuracy under Strategy A and the clean benchmark. Each group of bars corresponds to one embedding model. Green triangles ($\blacktriangle$) indicate the highest retrieval accuracy model at each chunk size; red triangles ($\blacktriangledown$) indicate the lowest.}
    \label{fig:Strategy_a_clean}
\end{figure}

As shown in Figure \ref{fig:Strategy_a_clean} on TeleEmbedBench-Clean, Per-corpus accuracy is marginally higher than on the main benchmark for every model, but the relative ordering is completely preserved. Qwen3 peaks at \textbf{0.662} on O-RAN (512 tokens) and \textbf{0.669} on 3GPP (2048 tokens), and Multi-MPNet remains the worst across all corpora and chunk sizes. Crucially, the performance drop incurred by comparing against the main benchmark is small for the LLM-based embedders and noticeably larger for the weaker sentence-transformers, confirming that Qwen3 and EGemma are more robust to incomplete and noisy queries, while models such as Multi-MPNet rely more heavily on well-formed input to achieve even their modest baseline scores.

\subsubsection{Strategy B (Macro-Averaged)}
\begin{figure}[t]
    \centering
    \includegraphics[width=\columnwidth]{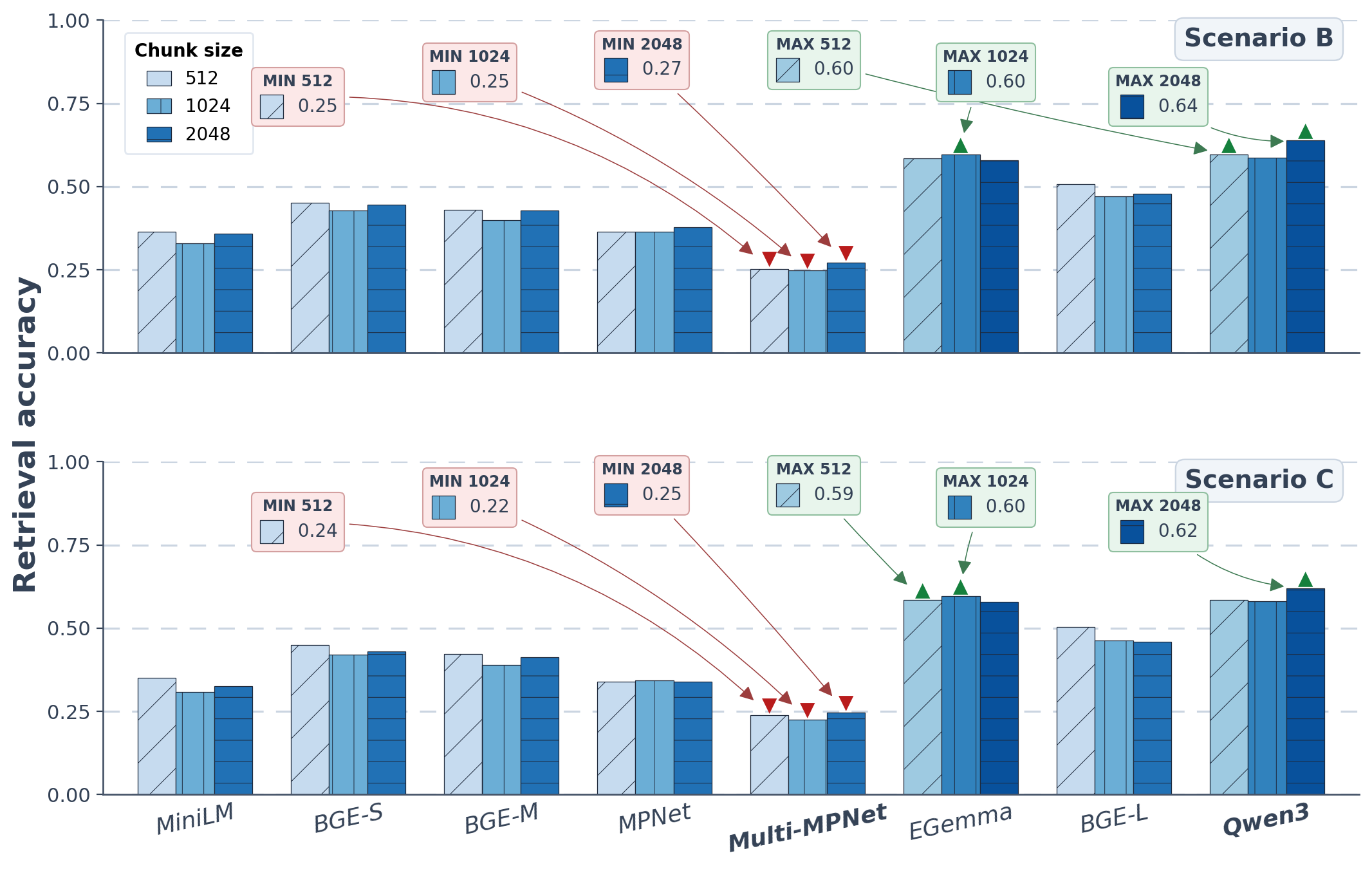}
    \caption{Retrieval accuracy under Strategy B and C for the clean benchmark. Each group of bars corresponds to one embedding model. Green triangles ($\blacktriangle$) indicate the highest retrieval accuracy model at each chunk size; red triangles ($\blacktriangledown$) indicate the lowest.}
    \label{fig:Strategy_bc_clean}
\end{figure}

\begin{figure}[t]
    \centering
    \includegraphics[width=\columnwidth]{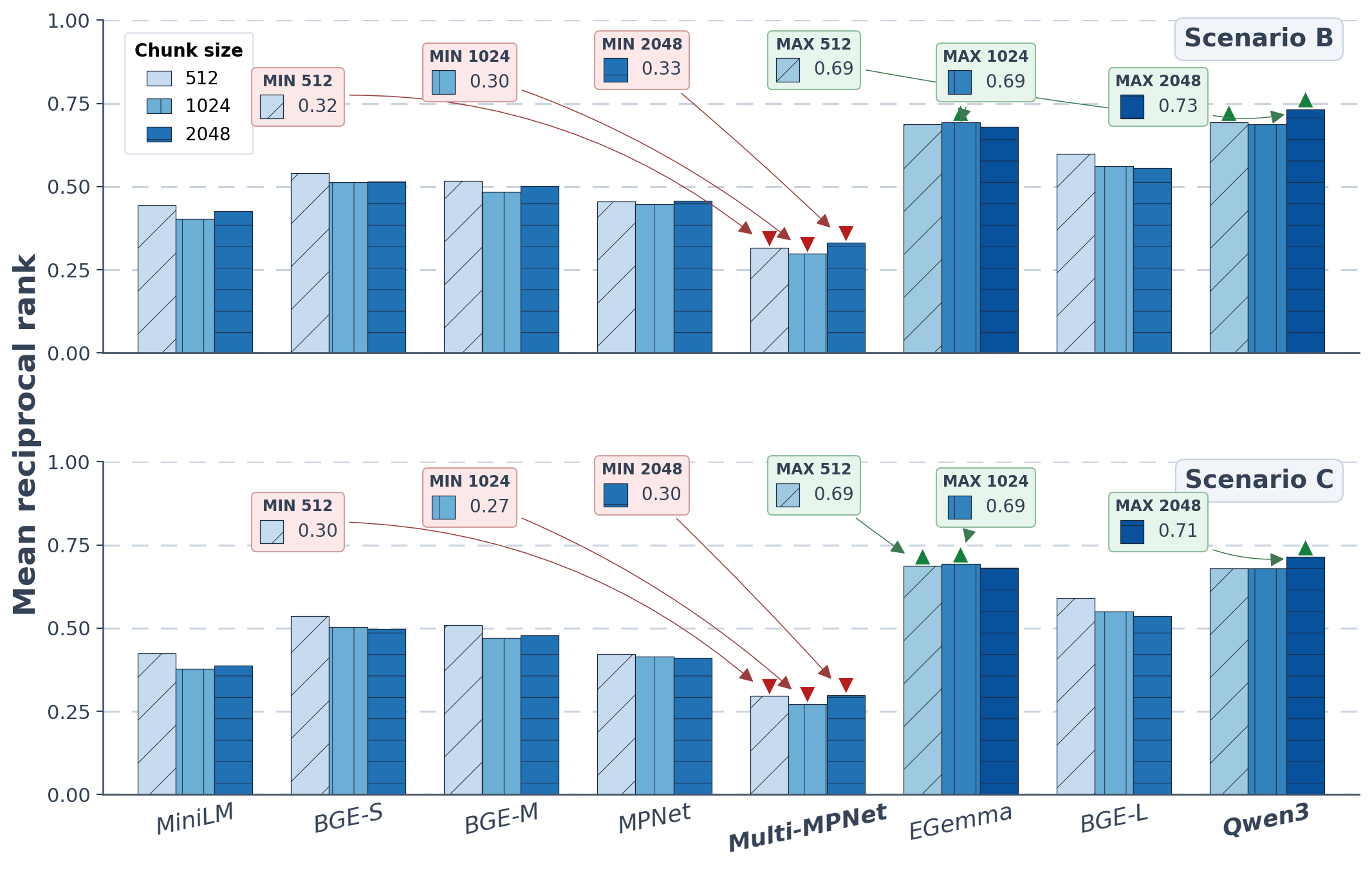}
    \caption{MMR under Strategy B and C for the clean benchmark. Each group of bars corresponds to one embedding model. Green triangles ($\blacktriangle$) indicate the highest retrieval accuracy model at each chunk size; red triangles ($\blacktriangledown$) indicate the lowest.}
    \label{fig:Strategy_bc_mrr_clean}
\end{figure}

The macro-averaged gap between TeleEmbedBench-Clean and the main benchmark quantifies query-noise robustness directly as shown in Figure \ref{fig:Strategy_bc_clean} and \ref{fig:Strategy_bc_mrr_clean}. At 2048 tokens, Qwen3 drops by only 0.039 when compared against the main benchmark (0.640~$\rightarrow$~0.601) and EGemma by the same margin (0.579~$\rightarrow$~0.540), whereas Multi-MPNet loses just 0.020 (0.271~$\rightarrow$~0.251), a smaller absolute drop but from a substantially lower ceiling, leaving it the worst model in both settings. The sentence-transformers in the middle tier (MiniLM, BGE-S, BGE-M, MPNet) suffer comparable or larger proportional losses than Qwen3 and EGemma, reinforcing that the LLM-based embedders extract sufficient semantic signal even from partial queries to maintain strong retrieval, while smaller architectures are more brittle to query degradation. MRR under Strategy B tells the same story: Qwen3's macro-averaged MRR falls from 0.733 to approximately 0.691 across chunk sizes when noise is added, whereas Multi-MPNet's already-low MRR of 0.332 erodes further, underscoring the robustness advantage of higher-capacity models.

\subsubsection{Strategy C (Merged FAISS)}

The robustness advantage of LLM-based embedders is most apparent under the unified FAISS index. On the clean benchmark, Qwen3 and EGemma incur minimal cross-domain penalties (e.g., Qwen3 at 2048: 0.640~$\rightarrow$~0.620, $-$0.020; EGemma at 1024: effectively no loss). The Strategy~C scores degrade more (e.g., EGemma at 2048: $-$0.046), yet they still outperform all sentence-transformers by a wide margin. Multi-MPNet, by contrast, suffers from both cross-domain interference \emph{and} query noise simultaneously, producing the largest combined drop and its overall minimum across all evaluation configurations.

\subsection{Teleco-specific Prompting}\label{sec:instruction}

\begin{table*}[t]
\centering
\caption{Retrieval accuracy and instruction impact by corpus on TeleEmbedBench and TeleEmbedBench-Clean with Qwen3}
\label{tab:combined-instruction-by-protocol}
\resizebox{\textwidth}{!}{%
\begin{tabular}{llcccccccccccccccc}
\toprule
\multirow{2}{*}{Dataset} & \multirow{2}{*}{Chunk} & \multicolumn{4}{c}{O-RAN} & \multicolumn{4}{c}{3GPP} & \multicolumn{4}{c}{srsRAN} & \multicolumn{4}{c}{Macro-avg.} \\
\cmidrule(lr){3-6} \cmidrule(lr){7-10} \cmidrule(lr){11-14} \cmidrule(lr){15-18}
 & & Base & Instr. & $\Delta$ & (\%) & Base & Instr. & $\Delta$ & (\%) & Base & Instr. & $\Delta$ & (\%) & Base & Instr. & $\Delta$ & (\%) \\
\midrule
\multirow{3}{*}{TeleEmbedBench}
 & 512  & 0.625 & 0.599 & $-0.026$ & $-4.16$ & 0.569 & 0.518 & $-0.051$ & $-8.96$ & 0.489 & 0.558 & $+0.069$ & $+14.11$ & 0.561 & 0.558 & $-0.003$ & $-0.48$ \\

 & 1024  & 0.603 & 0.599 & $-0.004$ & $-0.66$ & 0.610 & 0.541 & $-0.069$ & $-11.31$ & 0.489 & 0.561 & $+0.072$ & $+14.72$ & 0.567 & 0.567 & $-0.000$ & $-0.06$ \\

 & 2048  & 0.622 & 0.597 & $-0.025$ & $-4.02$ & 0.619 & 0.575 & $-0.044$ & $-7.11$ & 0.563 & 0.603 & $+0.04$ & $+7.1$ & 0.601 & 0.592 & $-0.01$ & $-1.61$ \\
\midrule
\multirow{3}{*}{TeleEmbedBench-Clean}
 & 512  & 0.662 & 0.643 & $-0.019$ & $-2.85$ & 0.609 & 0.549 & $-0.06$ & $-9.84$ & 0.489 & 0.558 & $+0.069$ & $+14.11$ & 0.587 & 0.584 & $-0.003$ & $-0.56$ \\

 & 1024  & 0.622 & 0.628 & $+0.006$ & $+0.98$ & 0.645 & 0.566 & $-0.079$ & $-12.3$ & 0.489 & 0.561 & $+0.072$ & $+14.72$ & 0.585 & 0.585 & $-0.000$ & $-0.07$ \\

 & 2048  & 0.649 & 0.629 & $-0.021$ & $-3.17$ & 0.669 & 0.625 & $-0.044$ & $-6.53$ & 0.563 & 0.603 & $+0.04$ & $+7.1$ & 0.627 & 0.619 & $-0.008$ & $-1.29$ \\
\bottomrule
\end{tabular}%
}
\end{table*}

Contrary to the conventional assumption that task-specific instructions strictly improve embedding alignment, our evaluation demonstrates that domain-specific prompting generally degrades retrieval performance for dense telecommunications specifications. As shown in Table \ref{tab:combined-instruction-by-protocol}, applying teleco-specific instructions to the Qwen3 base model yields a slight overall decline in macro-averaged accuracy across almost all configurations, such as a $-1.61\%$ drop for 2048 token chunks. This degradation is most pronounced within the heavily standardized textual corpora; the 3GPP dataset experiences severe accuracy drops across the board, peaking at a $-11.31\%$ decline for 1024 token chunks on the standard benchmark and a $-12.30\%$ drop on the Clean subset. 

The O-RAN corpus similarly suffers, with performance decreasing by up to $-4.16\%$ for 512 token chunks. While instruction prompting does provide a significant boost for retrieving raw C++ source code in the srsRAN corpus, improving accuracy by up to $+14.72\%$ by grounding the structurally bare code, the broader evidence clearly indicates that for natural language telecommunications standards, injecting verbose task instructions dilutes the model's semantic focus on critical, acronym-heavy keywords. Ultimately, these findings suggest that for standard text-based RAG pipelines operating on telecommunications prose, instruction prompting generally proves counterproductive and should be disabled to prevent query dilution.
\subsection{Downstream Analysis}\label{sec:downstream}

\begin{figure}[ht]
  \centering
  \includegraphics[width=\linewidth]{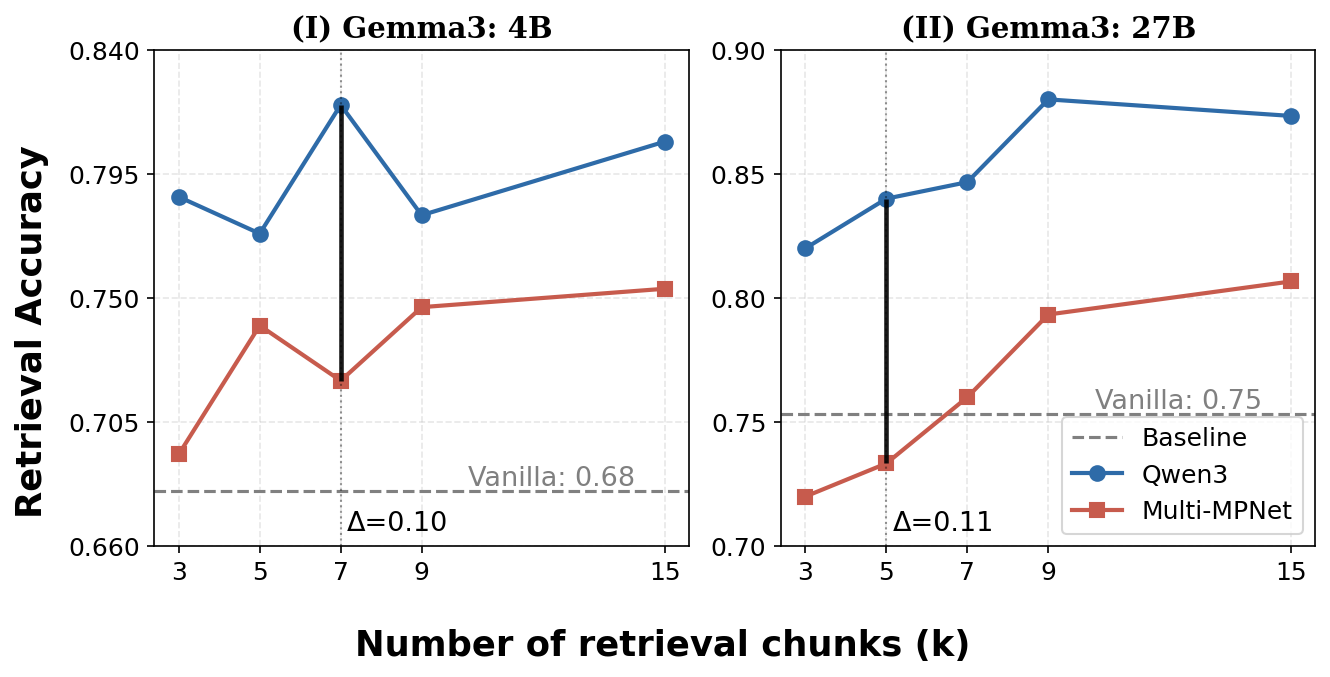}
  \caption{
  RAG ablation on ORANBench for two Gemma3 models: (I) Gemma3: 4B and (II) Gemma3: 27B}
  \label{fig:rag_ablation_4b_27b}
\end{figure}

Figure~\ref{fig:rag_ablation_4b_27b} presents a RAG ablation study on ORANBench\footnote{Here we use the ot-lite benchmark as provided by GSMA \cite{gsma2026} as it is the standard evaluation suite for teleco LLM models.} for two LLM models (Gemma3 4B and 27B). In each panel, we compare the downstream accuracy\footnote{Here, downstream accuracy is equivalent to the benchmark score. As ORANBench is an MCQA-based LLM benchmark, the task is to predict a correct option for a question; hence, the primary metric is \textit{accuracy}} obtained using a top-performing embedding model from our benchmark, Qwen3-Embedding-0.6B, against the lowest-ranked baseline, Multi-MPNet, with different numbers of retrieved chunks \(k \in \{3,5,7,9,15\}\) for the RAG inference. The dashed horizontal line denotes the \textit{Vanilla} accuracy of the corresponding Gemma3 model without any retrieved context, annotated on the right as \emph{Vanilla:}~\(p\). For each reader, we highlight the value of \(k\) at which the absolute performance gap \(\Delta = |{\rm Acc}_{\text{Qwen3}} - {\rm Acc}_{\text{Multi\text{-}MPNet}}|\) is maximal, annotating the resulting \(\Delta_{\max}\) on the x-axis. The results demonstrate a clear and consistent advantage for Qwen3 over Mult-MPNet across both reader sizes and all values of \(k\). While both RAG configurations significantly outperform the Vanilla baseline, confirming the necessity of retrieval for domain-specific telecommunications tasks, the quality of the retriever heavily dictates the ceiling of that downstream improvement. 
Notably, the downstream accuracy for the 4B model exhibits a markedly more erratic trend as $k$ increases, likely because its limited parameter capacity renders it highly susceptible to distraction from noise in the retrieved content. Conversely, for the 27B model, relying on the initially incorrect chunks retrieved by Multi-MPNet at lower k values yields the worst overall performance, demonstrating that highly capable models can be severely misdirected by flawed retrieved text.

\section{Discussion and Limitation}\label{limitations}
Excluding the aforementioned results for the embedding models we also observe a distinct correlation between the overall size of the corpus and the optimal chunk size for retrieval. Specifically, smaller corpuses such as the O-RAN specifications demonstrated superior performance when processed with a smaller chunk size of 512 tokens. Conversely, the massive volume and intricate structure of the 3GPP dataset heavily favored the maximum chunk size of 2048 tokens. We beleive that this phenomenon likely occurs because smaller specification sets contain highly condensed and localized technical requirements, making narrower context windows more effective at isolating relevant information without introducing excessive noise. In contrast, the expansive 3GPP standards involve extensive cross referencing and deeply nested procedural descriptions that require the broader context provided by the 2048 token limit to maintain semantic coherence during retrieval. Additionally we acknowledge the following limitations of our study:

\begin{itemize}[leftmargin=*]
    \item \noindent Our current evaluation framework is strictly constrained to embedding models containing fewer than one billion parameters. While this deliberate constraint ensures our benchmarking pipeline remains computationally accessible and realistic for deployment in resource-constrained environments, it inherently leaves the advanced performance capabilities and nuanced retrieval behaviors of massively scaled embedding models like the Qwen3-8B \cite{zhang2025qwen3} unexplored within the TeleEmbedBench.
    \item \noindent A notable limitation of our study is the constrained scope of our downstream end-to-end RAG evaluation, which is restricted to the O-RAN domain via the ORANBench13K dataset. While we acknowledge other domain-specific QA benchmarks such as srsRANBench \cite{gajjar2025oransight} and TeleQnA \cite{maatouk2025teleqna}, incorporating them presents a computationally intractable hurdle due to the strict version-alignment requirements inherent to RAG evaluations. For the downstream analysis to hold, the underlying retrieval database must perfectly mirror the corpus used to generate the benchmark questions. Consequently, evaluating against TeleQnA would necessitate constructing massive FAISS indices across the entirety of historical 3GPP releases, rather than the specific Release 19 documents targeted in TeleEmbedBench. Similarly, aligning with srsRANBench would demand re-indexing precise historic snapshots of the srsRAN codebase. Therefore, our downstream validation serves as a representative proxy limited to the O-RAN subset, even though our intrinsic retrieval metrics comprehensively cover all three domains.

    \item \noindent It is intuitive to believe that fine-tuning these embedding models on Telecom-specific texts should help it's performance, and we acknowledge the initial work performed by GSMA \cite{gsma2026, otel2026}. However, as limited information is available regarding how the models were trained and how the training data was collected, we include additional experiments in the appendix \ref{appx:ft}. We can observe how finetuning offers promising results, and we believe that obtaining a robust finetuning pipeline is an integral part of our future work.
\end{itemize}
\section{Conclusion}
In this work, we introduce \textit{TeleEmbedBench}, the first large-scale embedding benchmark tailored specifically for the telecommunications domain. This comprehensive benchmark encompasses O-RAN, 3GPP, and srsRAN. By addressing the critical limitations of general-purpose benchmarks like BEIR and MTEB, our evaluation of state-of-the-art representative eight models yields several actionable insights. First, LLM-based embedders such as Qwen3 and EmbeddingGemma consistently outperform the sentence-transformer family across all corpora and demonstrate superior robustness to noisy, incomplete queries. Second, we find that telecom-specific instruction prompting degrades retrieval on natural language specifications while improving it for structurally sparse source code. This finding emphasizes the need for domain-specific fine-tuning over generic prompting. Finally, downstream QA analysis on ORANBench validates that \textit{TeleEmbedBench} retrieval accuracy strongly correlates with end-to-end answer quality, proving its efficacy as a reliable evaluation framework for telecom RAG pipelines.


\bibliography{refs}

@article{zhou2024large,
  title={Large language model (llm) for telecommunications: A comprehensive survey on principles, key techniques, and opportunities},
  author={Zhou, Hao and Hu, Chengming and Yuan, Ye and Cui, Yufei and Jin, Yili and Chen, Can and Wu, Haolun and Yuan, Dun and Jiang, Li and Wu, Di and others},
  journal={IEEE Communications Surveys \& Tutorials},
  volume={27},
  number={3},
  pages={1955--2005},
  year={2024},
  publisher={IEEE}
}

@misc{bge_embedding,
      title={C-Pack: Packaged Resources To Advance General Chinese Embedding}, 
      author={Shitao Xiao and Zheng Liu and Peitian Zhang and Niklas Muennighoff},
      year={2023},
      eprint={2309.07597},
      archivePrefix={arXiv},
      primaryClass={cs.CL}
}

@article{tao2024llms,
  title={Llms are also effective embedding models: An in-depth overview},
  author={Tao, Chongyang and Shen, Tao and Gao, Shen and Zhang, Junshuo and Li, Zhen and Hua, Kai and Hu, Wenpeng and Tao, Zhengwei and Ma, Shuai},
  journal={arXiv preprint arXiv:2412.12591},
  year={2024}
}

@inproceedings{siino2024all,
  title={All-mpnet at semeval-2024 task 1: Application of mpnet for evaluating semantic textual relatedness},
  author={Siino, Marco},
  booktitle={Proceedings of the 18th International Workshop on Semantic Evaluation (SemEval-2024)},
  pages={379--384},
  year={2024}
}

@inproceedings{yin2024study,
  title={A study of sentence similarity based on the all-minilm-l6-v2 model with “same semantics, different structure” after fine tuning},
  author={Yin, Chen and Zhang, Zixuan},
  booktitle={2024 2nd International Conference on Image, Algorithms and Artificial Intelligence (ICIAAI 2024)},
  pages={677--684},
  year={2024},
  organization={Atlantis Press}
}

@inproceedings{wu2011optimizing,
  title={Optimizing mean reciprocal rank for person re-identification},
  author={Wu, Yang and Mukunoki, Masayuki and Funatomi, Takuya and Minoh, Michihiko and Lao, Shihong},
  booktitle={2011 8th IEEE International Conference on Advanced Video and Signal Based Surveillance (AVSS)},
  pages={408--413},
  year={2011},
  organization={IEEE}
}

@book{sidiropoulos2025improving,
  title={Improving the robustness and effectiveness of neural retrievers in noisy and low-resource settings},
  author={Sidiropoulos, Georgios},
  year={2025},
  publisher={Georgios Sidiropoulos}
}

@article{campos2023noise,
  title={Noise-robust dense retrieval via contrastive alignment post training},
  author={Campos, Daniel and Zhai, ChengXiang and Magnani, Alessandro},
  journal={arXiv preprint arXiv:2304.03401},
  year={2023}
}

@article{yang2025qwen3,
  title={Qwen3 technical report},
  author={Yang, An and Li, Anfeng and Yang, Baosong and Zhang, Beichen and Hui, Binyuan and Zheng, Bo and Yu, Bowen and Gao, Chang and Huang, Chengen and Lv, Chenxu and others},
  journal={arXiv preprint arXiv:2505.09388},
  year={2025}
}

@misc{gemmateam2025gemma3technicalreport,
      title={Gemma 3 Technical Report}, 
      author={Gemma Team and Aishwarya Kamath and Johan Ferret and Shreya Pathak and Nino Vieillard and Ramona Merhej and Sarah Perrin and Tatiana Matejovicova and Alexandre Ramé and Morgane Rivière and Louis Rouillard and Thomas Mesnard and Geoffrey Cideron and Jean-bastien Grill and Sabela Ramos and Edouard Yvinec and Michelle Casbon and Etienne Pot and Ivo Penchev and Gaël Liu and Francesco Visin and Kathleen Kenealy and Lucas Beyer and Xiaohai Zhai and Anton Tsitsulin and Robert Busa-Fekete and Alex Feng and Noveen Sachdeva and Benjamin Coleman and Yi Gao and Basil Mustafa and Iain Barr and Emilio Parisotto and David Tian and Matan Eyal and Colin Cherry and Jan-Thorsten Peter and Danila Sinopalnikov and Surya Bhupatiraju and Rishabh Agarwal and Mehran Kazemi and Dan Malkin and Ravin Kumar and David Vilar and Idan Brusilovsky and Jiaming Luo and Andreas Steiner and Abe Friesen and Abhanshu Sharma and Abheesht Sharma and Adi Mayrav Gilady and Adrian Goedeckemeyer and Alaa Saade and Alex Feng and Alexander Kolesnikov and Alexei Bendebury and Alvin Abdagic and Amit Vadi and András György and André Susano Pinto and Anil Das and Ankur Bapna and Antoine Miech and Antoine Yang and Antonia Paterson and Ashish Shenoy and Ayan Chakrabarti and Bilal Piot and Bo Wu and Bobak Shahriari and Bryce Petrini and Charlie Chen and Charline Le Lan and Christopher A. Choquette-Choo and CJ Carey and Cormac Brick and Daniel Deutsch and Danielle Eisenbud and Dee Cattle and Derek Cheng and Dimitris Paparas and Divyashree Shivakumar Sreepathihalli and Doug Reid and Dustin Tran and Dustin Zelle and Eric Noland and Erwin Huizenga and Eugene Kharitonov and Frederick Liu and Gagik Amirkhanyan and Glenn Cameron and Hadi Hashemi and Hanna Klimczak-Plucińska and Harman Singh and Harsh Mehta and Harshal Tushar Lehri and Hussein Hazimeh and Ian Ballantyne and Idan Szpektor and Ivan Nardini and Jean Pouget-Abadie and Jetha Chan and Joe Stanton and John Wieting and Jonathan Lai and Jordi Orbay and Joseph Fernandez and Josh Newlan and Ju-yeong Ji and Jyotinder Singh and Kat Black and Kathy Yu and Kevin Hui and Kiran Vodrahalli and Klaus Greff and Linhai Qiu and Marcella Valentine and Marina Coelho and Marvin Ritter and Matt Hoffman and Matthew Watson and Mayank Chaturvedi and Michael Moynihan and Min Ma and Nabila Babar and Natasha Noy and Nathan Byrd and Nick Roy and Nikola Momchev and Nilay Chauhan and Noveen Sachdeva and Oskar Bunyan and Pankil Botarda and Paul Caron and Paul Kishan Rubenstein and Phil Culliton and Philipp Schmid and Pier Giuseppe Sessa and Pingmei Xu and Piotr Stanczyk and Pouya Tafti and Rakesh Shivanna and Renjie Wu and Renke Pan and Reza Rokni and Rob Willoughby and Rohith Vallu and Ryan Mullins and Sammy Jerome and Sara Smoot and Sertan Girgin and Shariq Iqbal and Shashir Reddy and Shruti Sheth and Siim Põder and Sijal Bhatnagar and Sindhu Raghuram Panyam and Sivan Eiger and Susan Zhang and Tianqi Liu and Trevor Yacovone and Tyler Liechty and Uday Kalra and Utku Evci and Vedant Misra and Vincent Roseberry and Vlad Feinberg and Vlad Kolesnikov and Woohyun Han and Woosuk Kwon and Xi Chen and Yinlam Chow and Yuvein Zhu and Zichuan Wei and Zoltan Egyed and Victor Cotruta and Minh Giang and Phoebe Kirk and Anand Rao and Kat Black and Nabila Babar and Jessica Lo and Erica Moreira and Luiz Gustavo Martins and Omar Sanseviero and Lucas Gonzalez and Zach Gleicher and Tris Warkentin and Vahab Mirrokni and Evan Senter and Eli Collins and Joelle Barral and Zoubin Ghahramani and Raia Hadsell and Yossi Matias and D. Sculley and Slav Petrov and Noah Fiedel and Noam Shazeer and Oriol Vinyals and Jeff Dean and Demis Hassabis and Koray Kavukcuoglu and Clement Farabet and Elena Buchatskaya and Jean-Baptiste Alayrac and Rohan Anil and Dmitry and Lepikhin and Sebastian Borgeaud and Olivier Bachem and Armand Joulin and Alek Andreev and Cassidy Hardin and Robert Dadashi and Léonard Hussenot},
      year={2025},
      eprint={2503.19786},
      archivePrefix={arXiv},
      primaryClass={cs.CL},
      url={https://arxiv.org/abs/2503.19786}, 
}

@inproceedings{islam2025gpt,
  title={Gpt-4o: The cutting-edge advancement in multimodal llm},
  author={Islam, Raisa and Moushi, Owana Marzia},
  booktitle={Intelligent Computing-Proceedings of the Computing Conference},
  pages={47--60},
  year={2025},
  organization={Springer}
}

@article{mavroudis2024langchain,
  title={LangChain},
  author={Mavroudis, Vasilios},
  year={2024}
}

@inproceedings{barbosa2025open,
  title={Open-Source 5G RAN Platforms: A Dual Perspective on Performance and Capabilities},
  author={Barbosa, Maria and Gomes, Iasmin and Melo, Vin{\'\i}cius and Dias, Kelvin},
  booktitle={2025 Workshop on Communication Networks and Power Systems (WCNPS)},
  pages={1--7},
  year={2025},
  organization={IEEE}
}

@article{douze2025faiss,
  title={The faiss library},
  author={Douze, Matthijs and Guzhva, Alexandr and Deng, Chengqi and Johnson, Jeff and Szilvasy, Gergely and Mazar{\'e}, Pierre-Emmanuel and Lomeli, Maria and Hosseini, Lucas and J{\'e}gou, Herv{\'e}},
  journal={IEEE Transactions on Big Data},
  year={2025},
  publisher={IEEE}
}

@misc{otel2026,
  title={OTel: Open Telco AI Models},
  author={Tavakkoli, Farbod and Diamos, Gregory and Paulk, Roderic and Terrazas, Jorden},
  year={2026},
  url={https://huggingface.co/farbodtavakkoli}
}

@inproceedings{reimers2019sentence,
  title={Sentence-bert: Sentence embeddings using siamese bert-networks},
  author={Reimers, Nils and Gurevych, Iryna},
  booktitle={Proceedings of the 2019 conference on empirical methods in natural language processing and the 9th international joint conference on natural language processing (EMNLP-IJCNLP)},
  pages={3982--3992},
  year={2019}
}

@misc{devlin2019bertpretrainingdeepbidirectional,
      title={BERT: Pre-training of Deep Bidirectional Transformers for Language Understanding}, 
      author={Jacob Devlin and Ming-Wei Chang and Kenton Lee and Kristina Toutanova},
      year={2019},
      eprint={1810.04805},
      archivePrefix={arXiv},
      primaryClass={cs.CL},
      url={https://arxiv.org/abs/1810.04805}, 
}

@article{vera2025embeddinggemma,
  title={Embeddinggemma: Powerful and lightweight text representations},
  author={Vera, Henrique Schechter and Dua, Sahil and Zhang, Biao and Salz, Daniel and Mullins, Ryan and Panyam, Sindhu Raghuram and Smoot, Sara and Naim, Iftekhar and Zou, Joe and Chen, Feiyang and others},
  journal={arXiv preprint arXiv:2509.20354},
  year={2025}
}

@article{zhang2025qwen3,
  title={Qwen3 embedding: Advancing text embedding and reranking through foundation models},
  author={Zhang, Yanzhao and Li, Mingxin and Long, Dingkun and Zhang, Xin and Lin, Huan and Yang, Baosong and Xie, Pengjun and Yang, An and Liu, Dayiheng and Lin, Junyang and others},
  journal={arXiv preprint arXiv:2506.05176},
  year={2025}
}

@misc{gsma2026,
  title        = {Open Teleco},
  author       = {{GSMA}},
  year         = {2026},
  howpublished = {\url{https://github.com/gsma-labs/evals}}
}

@article{thakur2021beir,
  title={Beir: A heterogenous benchmark for zero-shot evaluation of information retrieval models},
  author={Thakur, Nandan and Reimers, Nils and R{\"u}ckl{\'e}, Andreas and Srivastava, Abhishek and Gurevych, Iryna},
  journal={arXiv preprint arXiv:2104.08663},
  year={2021}
}

@article{zhang2024scaling,
  title={When scaling meets llm finetuning: The effect of data, model and finetuning method},
  author={Zhang, Biao and Liu, Zhongtao and Cherry, Colin and Firat, Orhan},
  journal={arXiv preprint arXiv:2402.17193},
  year={2024}
}

@article{zou2025telecomgpt,
  title={Telecomgpt: A framework to build telecom-specific large language models},
  author={Zou, Hang and Zhao, Qiyang and Tian, Yu and Bariah, Lina and Bader, Faouzi and Lestable, Thierry and Debbah, Merouane},
  journal={IEEE Transactions on Machine Learning in Communications and Networking},
  year={2025},
  publisher={IEEE}
}

@article{yilma2025telecomrag,
  title={Telecomrag: Taming telecom standards with retrieval augmented generation and llms},
  author={Yilma, Girma M and Ayala-Romero, Jose A and Garcia-Saavedra, Andres and Costa-Perez, Xavier},
  journal={ACM SIGCOMM Computer Communication Review},
  volume={54},
  number={3},
  pages={18--23},
  year={2025},
  publisher={ACM New York, NY, USA}
}

@article{gajjar2025oransight,
  title={Oransight-2.0: Foundational llms for o-ran},
  author={Gajjar, Pranshav and Shah, Vijay K},
  journal={IEEE Transactions on Machine Learning in Communications and Networking},
  year={2025},
  publisher={IEEE}
}

@inproceedings{gajjar2025oran,
  title={Oran-bench-13k: An open source benchmark for assessing llms in open radio access networks},
  author={Gajjar, Pranshav and Shah, Vijay K},
  booktitle={2025 IEEE 22nd Consumer Communications \& Networking Conference (CCNC)},
  pages={1--4},
  year={2025},
  organization={IEEE}
}

@article{lotfi2025oran,
  title={ORAN-GUIDE: RAG-driven prompt learning for LLM-augmented reinforcement learning in o-ran network slicing},
  author={Lotfi, Fatemeh and Rajoli, Hossein and Afghah, Fatemeh},
  journal={arXiv preprint arXiv:2506.00576},
  year={2025}
}

@article{maatouk2025teleqna,
  title={Teleqna: A benchmark dataset to assess large language models telecommunications knowledge},
  author={Maatouk, Ali and Ayed, Fadhel and Piovesan, Nicola and De Domenico, Antonio and Debbah, Merouane and Luo, Zhi-Quan},
  journal={IEEE Network},
  year={2025},
  publisher={IEEE}
}

@inproceedings{muennighoff2023mteb,
  title={Mteb: Massive text embedding benchmark},
  author={Muennighoff, Niklas and Tazi, Nouamane and Magne, Lo{\"\i}c and Reimers, Nils},
  booktitle={Proceedings of the 17th Conference of the European Chapter of the Association for Computational Linguistics},
  pages={2014--2037},
  year={2023}
}

@inproceedings{ganiyu2025ai5gtest,
  title={AI5GTest: AI-Driven Specification-Aware Automated Testing and Validation of 5G O-RAN Components},
  author={Ganiyu, Abiodun and Gajjar, Pranshav and Shah, Vijay K},
  booktitle={18th ACM Conference on Security and Privacy in Wireless and Mobile Networks},
  pages={53--64},
  year={2025}
}

@article{chen2025each,
  title={Each to their own: Exploring the optimal embedding in rag},
  author={Chen, Shiting and Zhao, Zijian and Chen, Jinsong},
  journal={arXiv e-prints},
  pages={arXiv--2507},
  year={2025}
}
\bibliographystyle{ieeetr}
\begin{table*}[t]
\centering
\caption{Retrieval accuracy comparison between Qwen3 and OTel by corpus on TeleEmbedBench and TeleEmbedBench-Clean}
\label{tab:combined-qwen-otel-protocol}
\resizebox{\textwidth}{!}{%
\begin{tabular}{llcccccccccccccccc}
\toprule
\multirow{2}{*}{Dataset} & \multirow{2}{*}{Chunk} & \multicolumn{4}{c}{O-RAN} & \multicolumn{4}{c}{3GPP} & \multicolumn{4}{c}{srsRAN} & \multicolumn{4}{c}{Macro-avg.} \\
\cmidrule(lr){3-6} \cmidrule(lr){7-10} \cmidrule(lr){11-14} \cmidrule(lr){15-18}
& & Qwen3 & OTel & $\Delta$ & (\%) & Qwen3 & OTel & $\Delta$ & (\%) & Qwen3 & OTel & $\Delta$ & (\%) & Qwen3 & OTel & $\Delta$ & (\%) \\
\midrule
\multirow{3}{*}{TeleEmbedBench}
& 512  & 0.625 & 0.719 & $+0.094$ & $+15.04$ & 0.569 & 0.669 & $+0.100$ & $+17.57$ & 0.489 & 0.602 & $+0.113$ & $+23.11$ & 0.561 & 0.663 & $+0.102$ & $+18.24$ \\
& 1024 & 0.603 & 0.696 & $+0.093$ & $+15.42$ & 0.610 & 0.657 & $+0.047$ & $+7.70$ & 0.489 & 0.602 & $+0.113$ & $+23.11$ & 0.567 & 0.652 & $+0.084$ & $+14.86$ \\
& 2048 & 0.622 & 0.661 & $+0.039$ & $+6.27$ & 0.619 & 0.636 & $+0.017$ & $+2.75$ & 0.563 & 0.624 & $+0.061$ & $+10.83$ & 0.601 & 0.640 & $+0.039$ & $+6.49$ \\
\midrule
\multirow{3}{*}{TeleEmbedBench-Clean}
& 512  & 0.662 & 0.757 & $+0.095$ & $+14.27$ & 0.609 & 0.711 & $+0.102$ & $+16.73$ & 0.519 & 0.629 & $+0.111$ & $+21.30$ & 0.597 & 0.699 & $+0.102$ & $+17.14$ \\
& 1024 & 0.622 & 0.729 & $+0.107$ & $+17.25$ & 0.645 & 0.701 & $+0.056$ & $+8.67$ & 0.496 & 0.624 & $+0.128$ & $+25.89$ & 0.588 & 0.685 & $+0.097$ & $+16.54$ \\
& 2048 & 0.649 & 0.684 & $+0.035$ & $+5.36$ & 0.669 & 0.683 & $+0.015$ & $+2.17$ & 0.601 & 0.662 & $+0.061$ & $+10.22$ & 0.640 & 0.676 & $+0.037$ & $+5.77$ \\
\bottomrule
\end{tabular}%
}
\end{table*}

\newpage
\newpage

\appendix

\subsection{Prompt Templates}\label{prompts}
\begin{figure}[]
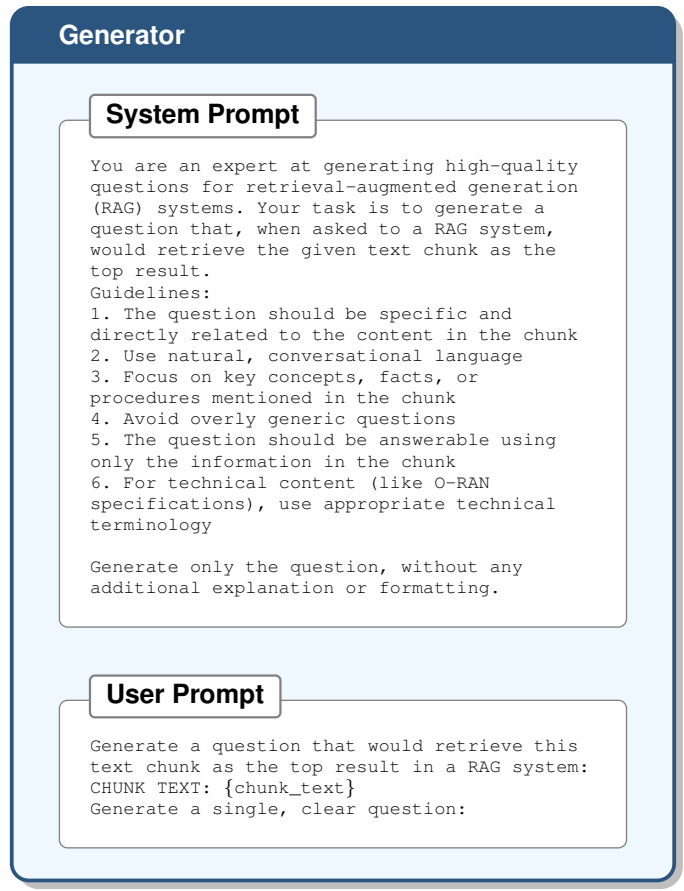

\centering
\begin{tcolorbox}[
  enhanced,
  colback=genbg,
  colframe=genframe,
  title={\sffamily\bfseries Generator},
  drop shadow,
  boxsep=2mm,
  arc=2mm,
  width=\columnwidth 
]
\begin{promptbox}{System Prompt}
\scriptsize
You are an expert at generating high-quality questions for retrieval-augmented generation (RAG) systems.
Your task is to generate a question that, when asked to a RAG system, would retrieve the given text chunk as the top result.
\\
Guidelines:\\
1. The question should be specific and directly related to the content in the chunk\\
2. Use natural, conversational language\\
3. Focus on key concepts, facts, or procedures mentioned in the chunk\\
4. Avoid overly generic questions\\
5. The question should be answerable using only the information in the chunk\\
6. For technical content (like O-RAN specifications), use appropriate technical terminology\\

Generate only the question, without any additional explanation or formatting.
\end{promptbox}

\vspace{2mm}
\begin{promptbox}{User Prompt}
\scriptsize

Generate a question that would retrieve this text chunk as the top result in a RAG system:

CHUNK TEXT:
\{chunk\_text\}

Generate a single, clear question:
\end{promptbox}

\end{tcolorbox}
\caption{System and User Prompts for the Question Generator model.}
\label{fig:prompts-generator}
\end{figure}
\begin{figure}[]
\centering
\begin{tcolorbox}[
  enhanced,
  colback=valbg,
  colframe=valframe,
  title={\sffamily\bfseries Validator},
  drop shadow,
  boxsep=2mm,
  arc=2mm,
  width=\columnwidth 
]
\begin{promptbox}{System Prompt}
\scriptsize
You are an expert validator for retrieval-augmented generation (RAG) benchmark datasets.
Your task is to validate whether a chunk-question pair is suitable for a RAG benchmark.

A valid pair must satisfy ALL of the following criteria:\\
1. **Text Content**: The chunk must be text-only (not a table, image description, or mostly non-text content)\\
2. **Question Quality**: The question must be clear, specific, and well-formed\\
3. **Relevance**: The question must be directly answerable using the information in the chunk\\
4. **Retrieval Suitability**: The question should be such that this chunk would be the top retrieval result]\\

Respond with a JSON object containing:\\
\{
  "is\_valid": true/false,
  "reasoning": "brief explanation of your decision",
  "issues": ["list of any issues found, empty if valid"]
\}
\end{promptbox}

\vspace{2mm}
\begin{promptbox}{User Prompt}
\scriptsize
Validate this chunk-question pair for a RAG benchmark:

CHUNK TEXT:
\{chunk\_text\}

QUESTION:
\{question\}

Evaluate the pair and respond with a JSON object as specified.
\end{promptbox}
\end{tcolorbox}
\caption{System and User Prompts for the Question Validator model.}
\label{fig:prompts-validator}
\end{figure}
\subsection{Fine Tuning}\label{appx:ft}

As detailed in Table \ref{tab:combined-qwen-otel-protocol}, our experiments evaluate the OTel-0.6B model \cite{otel2026}, which includes a domain-specifically fine-tuned variant of the baseline Qwen3-Embedding-0.6B. While the exact pretraining setup and the specific foundational documents utilized to train OTel remain undisclosed, the model demonstrates a consistent and significant retrieval performance advantage over the baseline across all evaluated corpora and chunk sizes. On the TeleEmbedBench-Clean dataset, OTel achieves a macro-average accuracy improvement of 17.14\% at a chunk size of 512 tokens, and maintains a 5.77\% lead at the broader 2048-token chunk size. These gains are particularly pronounced within the srsRAN corpus, where OTel yields a relative accuracy increase of up to 25.89\% for 1024-token chunks. Despite the lack of transparency regarding OTel's exact training regimen, these results strongly underscore that targeted fine-tuning is a highly promising avenue for adapting embedding models to the telecommunications domain, driving superior alignment with complex s like O-RAN and 3GPP.

\subsection{Validated Examples}\label{appx:val}
\begin{figure}[]
\begin{tcolorbox}[
  colback=gray!5, 
  colframe=blue!40!gray, 
  boxrule=1pt,
  sharp corners,
  title={\textbf{O-RAN} \hfill Chunk Size: \textbf{512}}
]
\scriptsize
\textbf{Chunk ID:} \\
\texttt{O-RAN.WG5.IOT.0-R003-v09.00\_241\_eb168075} 

\noindent\textcolor{blue!40!gray}{\rule{\linewidth}{0.5pt}}

\textbf{Source Document:} \\
\texttt{O-RAN.WG5.IOT.0-R003-v09.00.docx} 

\noindent\textcolor{blue!40!gray}{\rule{\linewidth}{0.5pt}}

\textbf{Chunk Text:}\\
One of the possible methods can be to make use of an O\&M command in the gNB-CU in order to initiate the Resource Status Reporting Initiation procedure to stop the measurement. Observe the Protocol Analyzer F1 logs\,\ldots,\ldots\ F1 logs recorded in the Protocol Analyzer are aligned with the message flows specified in the NR C-Plane profile specification Sections\,4.2.9.1.3\,/\,4.2.9.1.4.\,\ldots 

\noindent\textcolor{blue!40!gray}{\rule{\linewidth}{0.5pt}}

\textbf{Question:}\\
What method is used to stop a measurement in the gNB-CU, and what logs should be observed to verify the success of the Resource Status Reporting Initiation procedure according to the NR C-Plane profile specification? 

\noindent\textcolor{blue!40!gray}{\rule{\linewidth}{0.5pt}}

\textbf{Validation Reasoning:}\\
The question is clear, specific, and directly answerable from the provided chunk. The chunk explicitly describes the method (O\&M command) and the logs to observe (F1 logs aligned with the NR C-Plane profile specification). The question targets key information within the chunk, making it suitable for RAG evaluation.

\end{tcolorbox}
\caption{O-RAN chunk sample (Chunk Size: 512) from \texttt{O-RAN.WG5.IOT.0-R003-v09.00.docx}, illustrating the Resource Status Reporting Initiation procedure, associated question, and the validator's reasoning.}
\label{fig:oran-chunk-512}
\end{figure}

\begin{figure}[]
\begin{tcolorbox}[
  colback=gray!5, 
  colframe=blue!40!gray, 
  boxrule=1pt,
  sharp corners,
  title={\textbf{O-RAN} \hfill Chunk Size: \textbf{1024}}
]
\scriptsize
\textbf{Chunk ID:} \\
\texttt{O-RAN.WG6.O2DMS-INTERFACE-ETSI-NFV-PROFILE\\-R003-v06.00\_4\_c256fda9} 

\noindent\textcolor{blue!40!gray}{\rule{\linewidth}{0.5pt}}

\textbf{Source Document:} \\
\texttt{O-RAN.WG6.O2DMS-INTERFACE-ETSI-NFV-PROFILE-\\R003-v06.00.docx} 

\noindent\textcolor{blue!40!gray}{\rule{\linewidth}{0.5pt}}

\textbf{Chunk Text:}\\
\ldots for the orchestration and management operations through the O2dms ETSI NFV profile, it is assumed that descriptors and packaging of corresponding NF Deployments, and consequently of O-RAN NF, are on-boarded and available to the SMO and DMS \ldots The service interfaces and operations specified in the present document references the latest published Release 4 version of the ETSI GS NFV-SOL 003, which is the one used in the protocol and data model profiling in clause 3. As indicated in clause 2.2, the Release 4 version of ETSI GS NFV-SOL 003 adds support for OS container-based NF deployments, and some relevant features are also documented in clause 2.5. \ldots

\noindent\textcolor{blue!40!gray}{\rule{\linewidth}{0.5pt}}

\textbf{Question:}\\
Which ETSI NFV specifications are referenced?

\noindent\textcolor{blue!40!gray}{\rule{\linewidth}{0.5pt}}

\textbf{Validation Reasoning:}\\
The chunk is text-only and the question is clear and directly answerable from the provided text. The question targets a specific piece of information explicitly mentioned in the chunk. The chunk would likely be retrieved as a top result for this question.

\end{tcolorbox}
\caption{O-RAN chunk sample (Chunk Size: 1024) from \texttt{O-RAN.WG6.O2DMS-INTERFACE-ETSI-NFV-PROFILE-R003-v06.00.docx}, illustrating ETSI NFV profiling for O2dms interfaces, associated question, and the validator's reasoning.}
\label{fig:oran-chunk-1024}
\end{figure}

\begin{figure}[]
\begin{tcolorbox}[
  colback=gray!5, 
  colframe=blue!40!gray, 
  boxrule=1pt,
  sharp corners,
  title={\textbf{O-RAN} \hfill Chunk Size: \textbf{2048}}
]
\scriptsize
\textbf{Chunk ID:} \\
\texttt{O-RAN.WG9.XTRP-TST.0-R003-v03.00\_25\_64cc1f39} 

\noindent\textcolor{blue!40!gray}{\rule{\linewidth}{0.5pt}}

\textbf{Source Document:} \\
\texttt{O-RAN.WG9.XTRP-TST.0-R003-v03.00.pdf} 

\noindent\textcolor{blue!40!gray}{\rule{\linewidth}{0.5pt}}

\textbf{Chunk Text:}\\
\,\ldots\ Test ID Ethernet.IPv6.01 Classification Functionality \newline
Test Title Native support of IPv6 connectivity\,\ldots\ Test Procedure: \newline
1. Define a native IPv6 connectivity in TNEs. \newline
2. Perform connectivity between O-RU, O-DU and O-CU natively in IPv6 (if eCPRI is directly encapsulated in Eth, then this would apply solely to midhaul traffic). \newline
3. The test should be at least 120 secs in duration.\,\ldots\ Pass/Fail Criteria: Compare the results obtained for latency, jitter and throughput for the scenarios with and without background traffic.\,\ldots\ 

\noindent\textcolor{blue!40!gray}{\rule{\linewidth}{0.5pt}}

\textbf{Question:}\\
What are the test procedures and pass/fail criteria for validating native IPv6 connectivity in O-RU, O-DU, and O-CU within an O-RAN system, including comparisons of latency, jitter, and throughput with and without background traffic?

\noindent\textcolor{blue!40!gray}{\rule{\linewidth}{0.5pt}}

\textbf{Validation Reasoning:}\\
The question is directly answerable from the provided chunk. The chunk details the test procedures (steps 1-3) and pass/fail criteria (comparing latency, jitter, and throughput with and without background traffic) for IPv6 connectivity tests. The question is clear and specific, and the chunk is a relevant source for the answer. The chunk would likely be retrieved as a top result for this question.

\end{tcolorbox}
\caption{O-RAN chunk sample (Chunk Size: 2048) from \texttt{O-RAN.WG9.XTRP-TST.0-R003-v03.00.pdf}, illustrating native IPv6 connectivity test procedures and pass/fail criteria, associated question, and the validator's reasoning.}
\label{fig:oran-chunk-2048}
\end{figure}

\begin{figure}[]
\begin{tcolorbox}[
  colback=gray!5,
  colframe=violet!35!gray,
  boxrule=1pt,
  sharp corners,
  title={srsRAN \hfill Chunk Size: 512}
]
\scriptsize
Chunk ID: \\
\texttt{s1ap\_2128\_37\_f4d01d1c}

\noindent\textcolor{violet!35!gray}{\rule{\linewidth}{0.5pt}}

Source Document: \\
\texttt{s1ap.h}

\noindent\textcolor{violet!35!gray}{\rule{\linewidth}{0.5pt}}

Chunk Text:\\
\texttt{using completed\_cellin\_eai\_l = dyn\_array<completed\_cellin\_eai\_item\_s>;} \ldots\
\texttt{struct cell\_id\_broadcast\_item\_s \{ bool ext = false; bool ie\_exts\_present = false; eutran\_cgi\_s ecgi; \ldots\ SRSASN\_CODE pack(bit\_ref\& bref) const; \ldots\ \};} \ldots\
\texttt{struct emergency\_area\_id\_broadcast\_item\_s \{ bool ext = false; \ldots\ fixed\_octstring<3, true> emergency\_area\_id; completed\_cellin\_eai\_l completed\_cellin\_eai; \ldots\ void to\_json(json\_writer\& j) const; \};}\ \ldots

\noindent\textcolor{violet!35!gray}{\rule{\linewidth}{0.5pt}}

Question:\\
What are the key components and structure of the \texttt{emergency\_area\_id\_broadcast\_item\_s} data structure as defined in the provided text?

\noindent\textcolor{violet!35!gray}{\rule{\linewidth}{0.5pt}}

Validation Reasoning:\\
The chunk provides a detailed definition of the \texttt{emergency\_area\_id\_broadcast\_item\_s} structure, including its members and methods. The question directly asks for the components and structure of this data structure, making it highly relevant and answerable from the chunk. The question is clear and well-formed. This chunk would likely be a top retrieval result for this question.

\end{tcolorbox}
\caption{srsRAN chunk sample (Chunk Size: 512) from \texttt{s1ap.h}, illustrating the \texttt{emergency\_area\_id\_broadcast\_item\_s} structure definition, associated question, and the validator's reasoning.}
\label{fig:srsran-chunk-512}
\end{figure}

\begin{figure}[]
\begin{tcolorbox}[
  colback=gray!5,
  colframe=violet!35!gray,
  boxrule=1pt,
  sharp corners,
  title={srsRAN \hfill Chunk Size: 1024}
]
\scriptsize
Chunk ID: \\
\texttt{e2ap\_7b8a\_152\_b9e17cf0}

\noindent\textcolor{violet!35!gray}{\rule{\linewidth}{0.5pt}}

Source Document: \\
\texttt{e2ap.cpp}

\noindent\textcolor{violet!35!gray}{\rule{\linewidth}{0.5pt}}

Chunk Text:\\
\texttt{void unsuccessful\_outcome\_s::to\_json(json\_writer\& j) const \{ j.start\_obj(); j.write\_int(``procedureCode'', proc\_code); \ldots\ j.end\_obj(); \}} \ldots\
\texttt{bool unsuccessful\_outcome\_s::load\_info\_obj(const uint16\_t\& proc\_code\_) \{ \ldots\ proc\_code = proc\_code\_; crit = e2\_ap\_elem\_procs\_o::get\_crit(proc\_code); \ldots\ \}} \ldots\
\texttt{void e2\_ap\_pdu\_c::set(types::options e) \{ destroy\_(); type\_ = e; switch (type\_) \{ case types::init\_msg: c.init<init\_msg\_s>(); break; \ldots\ \} \}} \ldots\
\texttt{e2\_ap\_pdu\_c\& e2\_ap\_pdu\_c::operator=(const e2\_ap\_pdu\_c\& other) \{ if (this == \&other) \{ return *this; \} set(other.type()); \ldots\ \}}

\noindent\textcolor{violet!35!gray}{\rule{\linewidth}{0.5pt}}

Question:\\
How does the \texttt{to\_json} method of the \texttt{unsuccessful\_outcome\_s} class serialize its data members into a JSON object?

\noindent\textcolor{violet!35!gray}{\rule{\linewidth}{0.5pt}}

Validation Reasoning:\\
The chunk provides the exact code for the \texttt{to\_json} method of the \texttt{unsuccessful\_outcome\_s} class, directly answering the question. The question is clear and specific, and the chunk is relevant and suitable for retrieval.

\end{tcolorbox}
\caption{srsRAN chunk sample (Chunk Size: 1024) from \texttt{e2ap.cpp}, illustrating the \texttt{to\_json} serialization and PDU choice handling logic, associated question, and the validator's reasoning.}
\label{fig:srsran-chunk-1024}
\end{figure}

\begin{figure}[]
\begin{tcolorbox}[
  colback=gray!5,
  colframe=violet!35!gray,
  boxrule=1pt,
  sharp corners,
  title={srsRAN \hfill Chunk Size: 2048}
]
\scriptsize
Chunk ID: \\
\texttt{common\_ext\_8a5f\_12\_9614004a}

\noindent\textcolor{violet!35!gray}{\rule{\linewidth}{0.5pt}}

Source Document: \\
\texttt{common\_ext.cc}

\noindent\textcolor{violet!35!gray}{\rule{\linewidth}{0.5pt}}

Chunk Text:\\
\texttt{SRSASN\_CODE sl\_v2x\_inter\_freq\_ue\_cfg} \ldots\
\texttt{if (pci\_list\_r14\_present) \{ HANDLE\_CODE(unpack\_dyn\_seq\_of(pci\_list\_r14, bref, 1, 16, integer\_packer<uint16\_t>(0, 503))); \}} \ldots\
\texttt{return SRSASN\_SUCCESS; \}} \ldots\
\texttt{void sl\_v2x\_inter\_freq\_ue\_cfg\_r14\_s::to\_json(json\_writer\& j) const \{ j.start\_obj(); \ldots\ j.end\_obj(); \}} \ldots\
\texttt{SRSASN\_CODE cell\_sel\_info\_nfreq\_r13\_s::pack(bit\_ref\& bref) const \{ \ldots\ \}} \ldots\
\texttt{void sl\_allowed\_carrier\_freq\_list\_r15\_s::to\_json \ldots\ j.end\_obj(); \}}

\noindent\textcolor{violet!35!gray}{\rule{\linewidth}{0.5pt}}

Question:\\
What is the implementation of the \texttt{unpack()} method for the \texttt{sl\_v2x\_inter\_freq\_ue\_cfg\_r14\_s} struct?

\noindent\textcolor{violet!35!gray}{\rule{\linewidth}{0.5pt}}

Validation Reasoning:\\
The chunk contains the implementation of the \texttt{unpack()} method for the \texttt{sl\_v2x\_inter\_freq\_ue\_cfg\_r14\_s} struct. The question directly asks for the implementation of this method, making the question highly relevant to the chunk's content. The question is clear and specific. This chunk would likely be retrieved as a top result for this question.

\end{tcolorbox}
\caption{srsRAN chunk sample (Chunk Size: 2048) from \texttt{common\_ext.cc}, illustrating V2X inter-frequency UE configuration unpacking and JSON serialization, associated question, and the validator's reasoning.}
\label{fig:srsran-chunk-2048}
\end{figure}

\begin{figure}[]
\begin{tcolorbox}[
  colback=gray!5, 
  colframe=green!30!gray, 
  boxrule=1pt,
  sharp corners,
  title={\textbf{3GPP} \hfill Chunk Size: \textbf{512}}
]
\scriptsize
\textbf{Chunk ID:} \\
\texttt{43055-j00\_3\_55778a8e} 

\noindent\textcolor{green!30!gray}{\rule{\linewidth}{0.5pt}}

\textbf{Source Document:} \\
\texttt{43055-j00.docx} 

\noindent\textcolor{green!30!gray}{\rule{\linewidth}{0.5pt}}

\textbf{Chunk Text:}\\
\ldots\ the second digit is incremented for all changes of substance, i.e. technical enhancements, corrections, updates, etc.\,\ldots\ The definition of GPRS class A mode of operation in Release 97 assumes a total independence between the CS and PS domains. Thus the direct implementation of the existent standards for class A would result in mobile stations that are required to operate in two different frequencies either in the same timeslot, in timeslots n and n + 3 or their adjacent ones.\,\ldots\ operators have expressed their need for this type of mobiles, since they want to offer services that demand the simultaneous existence of a CS connection and a PS session.\,\ldots\ A constant aim throughout this document is to reuse the existing functionality when possible, in order to minimise the impact on current implementations.\,\ldots

\noindent\textcolor{green!30!gray}{\rule{\linewidth}{0.5pt}}

\textbf{Question:}\\
What technical challenge arises from the definition of GPRS class A mode in Release 97 due to the independence between CS and PS domains?

\noindent\textcolor{green!30!gray}{\rule{\linewidth}{0.5pt}}

\textbf{Validation Reasoning:}\\
The question is directly answerable from the provided chunk, specifically the paragraph discussing the definition of GPRS class A mode and the resulting architectural complications. The question is clear, specific, and well-formed. The chunk would likely be retrieved as a top result for this question.

\end{tcolorbox}
\caption{3GPP chunk sample (Chunk Size: 512) from \texttt{43055-j00.docx}, illustrating GPRS class A mode architectural challenges, associated question, and the validator's reasoning.}
\label{fig:3gpp-chunk-512}
\end{figure}

\begin{figure}[]
\begin{tcolorbox}[
  colback=gray!5, 
  colframe=green!30!gray, 
  boxrule=1pt,
  sharp corners,
  title={\textbf{3GPP} \hfill Chunk Size: \textbf{1024}}
]
\scriptsize
\textbf{Chunk ID:} \\
\texttt{23281-j90\_35\_7c741ea3} 

\noindent\textcolor{green!30!gray}{\rule{\linewidth}{0.5pt}}

\textbf{Source Document:} \\
\texttt{23281-j90.docx} 

\noindent\textcolor{green!30!gray}{\rule{\linewidth}{0.5pt}}

\textbf{Chunk Text:}\\
\ldots\ MCVideo server checks whether the MCVideo user at MCVideo client 1 is authorized to initiate the private call, and that MCVideo user at MCVideo client 2 is authorized to receive the private call. MCVideo server verifies whether the provided functional alias, if present, can be used and has been activated for the user.\,\ldots\ MCVideo client 1 and MCVideo client 2 have successfully established media plane and transmission control for communication and both users can transmit media.\,\ldots

\noindent\textcolor{green!30!gray}{\rule{\linewidth}{0.5pt}}

\textbf{Question:}\\
What steps are taken by the MCVideo server when processing a private call request that uses a functional alias instead of an MCVideo ID, including authorization checks, alias resolution, and security association?

\noindent\textcolor{green!30!gray}{\rule{\linewidth}{0.5pt}}

\textbf{Validation Reasoning:}\\
The question directly addresses information present in the chunk. The chunk details the steps the MCVideo server takes when a functional alias is used, including authorization, resolution, and security association setup. The question is clear, specific, and well-formed. This chunk would likely be retrieved as a top result for this question.

\end{tcolorbox}
\caption{3GPP chunk sample (Chunk Size: 1024) from \texttt{23281-j90.docx}, illustrating MCVideo server private call handling with functional alias resolution, associated question, and the validator's reasoning.}
\label{fig:3gpp-chunk-1024}
\end{figure}

\begin{figure}[t!]
\begin{tcolorbox}[
  colback=gray!5, 
  colframe=green!30!gray, 
  boxrule=1pt,
  sharp corners,
  title={\textbf{3GPP} \hfill Chunk Size: \textbf{2048}}
]
\scriptsize
\textbf{Chunk ID:} \\
\texttt{24554-j40\_54\_41f9f055} 

\noindent\textcolor{green!30!gray}{\rule{\linewidth}{0.5pt}}

\textbf{Source Document:} \\
\texttt{24554-j40.docx} 

\noindent\textcolor{green!30!gray}{\rule{\linewidth}{0.5pt}}

\textbf{Chunk Text:}\\
\ldots\ the application layer group ID parameter of the PROSE PC5 DISCOVERY message for group member discovery response is the same as the application layer group ID parameter of the PROSE PC5 DISCOVERY message for group member discovery solicitation\,\ldots\ When the UE is triggered by an upper layer application to stop soliciting proximity of other UEs in a discovery group, or when the UE stops being \,\ldots\ Upon reception of a PROSE PC5 DISCOVERY message for group member discovery solicitation, for the application layer group ID of the discovery group which the UE is configured to respond for, the UE shall use the associated DUSK\,\ldots\ to unscramble the PROSE PC5 DISCOVERY message.\,\ldots

\noindent\textcolor{green!30!gray}{\rule{\linewidth}{0.5pt}}

\textbf{Question:}\\
Under what conditions does a UE consider another UE in a discovery group as discovered based on the application layer group ID and target information parameters in the PROSE PC5 DISCOVERY message?

\noindent\textcolor{green!30!gray}{\rule{\linewidth}{0.5pt}}

\textbf{Validation Reasoning:}\\
The question directly addresses information present in the chunk. The chunk explicitly outlines the conditions under which a UE considers another UE discovered, referencing application layer group ID and target information. The question is clear, specific, and well-formed. The chunk would likely be a top retrieval result for this question. The text is also entirely text-based.

\end{tcolorbox}
\caption{3GPP chunk sample (Chunk Size: 2048) from \texttt{24554-j40.docx}, illustrating 5G ProSe group member discovery conditions and discoveree UE procedures, associated question, and the validator's reasoning.}
\label{fig:3gpp-chunk-2048}
\end{figure}

\end{document}